\documentclass[journal]{IEEEtran}
\usepackage[left=0.625in,right=0.625in,top=0.75in,bottom=1in]{geometry}
\usepackage{ifpdf}
\usepackage{cite}
\usepackage{array}
\usepackage{dblfloatfix}
\usepackage{url}
\usepackage{amsmath}
\usepackage{ragged2e}
\usepackage{graphicx}
\usepackage{verbatim}
\usepackage{float}
\usepackage{lipsum}
\usepackage{multicol, blindtext}
\usepackage{caption}
\usepackage{subcaption}
\usepackage{xcolor}
\definecolor{mypink2}{RGB}{0, 0, 255}
\definecolor{green}{RGB}{0, 128, 0}
\usepackage{multirow}
\usepackage{array}
\usepackage{tabularx}
\usepackage[acronym]{glossaries}
\usepackage[linesnumbered]{algorithm2e}
\usepackage{booktabs}
\usepackage{pifont}

\usepackage[most]{tcolorbox}
\tcbuselibrary{breakable}
\usepackage{colortbl}
\usepackage{soul}
\sethlcolor{yellow!20}
\usepackage{hyperref}
\usepackage{amssymb}

\begin{document}
\title{\fontsize{14pt}{14pt}\selectfont On the Use of AI-Driven Immersive Digital Technologies for Designing and Operating UAVs}

\author{Yousef~Emami,~\IEEEmembership{Senior Member,~IEEE,}         Mohammadhossein~Homaei,~\IEEEmembership{Senior Member,~IEEE,}         and~Miguel~Gutiérrez~Gaitán,~\IEEEmembership{Senior Member,~IEEE,}~and 
Mohammad~Shojafar,~\IEEEmembership{Senior Member,~IEEE}

\thanks{Copyright (c) 2026 IEEE. Personal use of this material is permitted. However, permission to use this material for any other purposes must be obtained from the IEEE by sending a request to pubs-permissions@ieee.org.}             
}
\maketitle

\begin{abstract}
Uncrewed Aerial Vehicles (UAVs) offer agile, cost-effective, and efficient solutions for communication relay networks. However, their modeling and control are challenging, and the mismatch between simulations and actual conditions limits real-world deployment, while maintaining adequate situational awareness remains essential for safe operation. Several studies have proposed integrating UAV operations with immersive digital technologies, such as Digital Twin (DT) and Extended Reality (XR), to overcome these challenges. This paper provides a comprehensive overview of the latest research and developments involving immersive digital technologies for UAVs. We explore the use of Machine Learning (ML) techniques, particularly Deep Reinforcement Learning (DRL), to improve the capabilities of DT for UAV systems, and present a case study of a DT-driven DRL pipeline that couples bidirectional physical-digital synchronization with online recursive least-squares channel calibration for UAV resource allocation. We further present a second case study in which a diffusion-augmented digital twin, kept statistically faithful to the physical swarm by the same online calibration loop, drives multi-UAV velocity coordination. We identify and discuss key research gaps, and propose countermeasures based on Generative AI (GAI), emphasizing the significant role of AI in advancing DT technology for UAVs. Furthermore, we review and discuss how the XR technology can transform UAV operations with the support of GAI, and examine its practical challenges. Finally, we propose future research directions to further develop the application of immersive digital technologies for UAV operation.
\end{abstract}

\begin{IEEEkeywords}
 -Uncrewed Aerial Vehicles, Digital Twin, Extended Reality, Virtual Reality, Augmented Reality, Mixed Reality, Machine Learning, Reinforcement Learning, Generative AI.
\end{IEEEkeywords}
\IEEEpeerreviewmaketitle

\section{Introduction}
\begin{table}[h]
    \centering
    \caption{List of Acronyms}
    \label{tab:acronyms}
    \begin{tabular}{|p{1.25cm}|p{6.8cm}|}
        \hline
        \multicolumn{1}{|c|}{\textbf{Acronym}}& \multicolumn{1}{|c|}{\textbf{Definition}}  \\
        
          \hline
        A3C    & Advantage Actor-Critic \\
        AI     & Artificial Intelligence \\
        AR     & Augmented Reality \\
        AR-HUD & Head-Up Display \\
        AVs    & Autonomous Vehicles \\
        BCDDPG & Behavior-Coupling Deep Deterministic Policy Gradient \\
        CPS    & Cyber-Physical Systems \\
        DDPG   & Deep Deterministic Policy Gradients \\
        DL     & Deep Learning \\
        DRL    & Deep Reinforcement Learning \\
        DQN    & Deep-Q-Network \\
        DT     & Digital Twin \\
        DTN    & Digital Twin Network \\
        eMBB   & Enhanced Mobile Broadband\\
        FDI    & False Data Injection\\
        GAI    & Generative AI \\
        GAN    & Generative Adversarial Network \\
        IoT    & Internet of Things \\
        ITS    & Intelligent Transportation Systems \\
        ISCC    & Integrated Sensing, Communication, and Computation\\
        LSTM   & Long Short-Term Memory \\
        MDP    & Markov Decision Process \\
        ML     & Machine Learning \\
        MR     & Mixed Reality \\
        MDP    & Markov Decision Process\\
        NR     & New Radio\\
        QoS    & Quality of Service \\ 
        RL     & Reinforcement Learning \\
        TAM    & Timed Automaton Machine \\
        UAV    & Uncrewed Aerial Vehicle \\
        URLLC  & Ultra-Reliable and Low Latency Communication\\
        VR     & Virtual Reality \\
        XR     & Extended Reality \\
        RLS     & Recursive Least-Squares  \\
        AoI     & Age of Information  \\
        MEC     & Mobile Edge Computing  \\
        mMTC    & Massive Machine-Type Communications\\
        HMD     & Head-Mounted Displays  \\
        FPV     & First Person View   \\
        TPV     & Third Person View   \\

        \hline
    \end{tabular}
\end{table}
Uncrewed Aerial Vehicles (UAVs), commonly known as drones, have made considerable progress in recent decades\cite{9920736,9793853}. The short line-of-sight communication links among UAVs and between them 
and their users, together with the controllable mobility, 
cost-effectiveness and agility, have led to their widespread use as aerial base stations \cite{8601408}, aerial relays\cite{9128998}, and aerial data collectors \cite{8854903}. Their adoption continues to grow across numerous civil and commercial sectors, including public safety \cite{10246260}, energy \cite{9276537}, agriculture \cite{9316211}, environmental monitoring \cite{9797309}, among others \cite{9917491}. In parallel with the development of UAVs, immersive digital technologies such as Digital Twin (DT) and Extended Reality (XR) have proven to be transformative forces in wireless systems \cite{9854866,8821836}.
A related list of acronyms is provided in Table~\ref{tab:acronyms}. 

\begin{table*}[!t]
\caption{Comparison of existing surveys with the proposed survey.}
\label{tab:survey_comparison}
\centering
\footnotesize
\renewcommand{\arraystretch}{1.2}
\begin{tabularx}{\textwidth}{|p{1cm}|p{1.8cm}|c|c|c|c|c|X|X|}
\hline
\textbf{Ref.} &
\textbf{Domain} &
\textbf{DT} &
\textbf{XR} &
\textbf{AI/ML} &
\textbf{DRL} &
\textbf{UAV} &
\textbf{Main Contribution} &
\textbf{Difference from This Survey} \\
\hline

\cite{robotics9020021} & Robotics & No & AR & No & No & No &
Reviews AR applications in robotics, including medical robotics, path planning, control, and multi-agent systems. &
Limited to AR in robotics; does not address DT, UAVs, AI-enhanced DTs, or immersive UAV operations.\\
\hline

\cite{su14020601} & ITS & Yes & No & Partial & No & No &
Reviews Digital Twin technology for Electric Vehicles and Autonomous Vehicles. &
Focuses on ITS rather than UAVs; excludes XR, DRL, and immersive technologies.\\
\hline

\cite{app13105871} & ITS & Yes & No & Partial & No & No &
Surveys DT applications for EVs and AVs. &
Vehicle-centric survey without UAV, XR, or AI-enabled DT integration.\\
\hline

\cite{9543560} & Autonomous Vehicles & No & VR & No & No & No &
Reviews VR-based studies on pedestrian--AV interaction. &
Focuses only on VR interaction; no DT, UAV, or AI integration.\\
\hline

\cite{kettle_safetyreview} & Automated Driving & No & AR & No & No & No &
Systematic review of AR visualizations for in-vehicle driver communication. &
Restricted to AR interfaces for drivers; does not consider UAVs or DT.\\
\hline

\cite{riegler2021systematic} & Automated Driving & No & VR & No & No & No &
Reviews VR research in automated driving. &
Limited to VR applications; excludes DT, UAVs, and AI-enhanced immersive systems.\\
\hline

\cite{GUO2022} & Internet of Vehicles & Yes & No & Partial & No & No &
Reviews Internet of Vehicles and Digital Twin technologies. &
Network-centric perspective without UAVs or XR technologies.\\
\hline

\cite{anandaraj2024survey} & UAVs & Yes & No & Yes & Partial & Yes &
Surveys AI and DT integration for UAV intelligence, learning models, and data processing. &
Covers AI and DT but omits XR, sim-to-real transfer, Generative AI, and immersive UAV operations.\\
\hline

\cite{10556896} & UAVs & Yes & No & Partial & No & Yes &
Reviews implementation tools and frameworks for UAV Digital Twins. &
Implementation-oriented; lacks XR, AI algorithms, and comprehensive UAV applications.\\
\hline

\cite{10570372} & DT Networks & Yes & No & Yes & Partial & Partial &
Reviews Digital Twin Networks with emphasis on ML and DL. &
Network-focused rather than UAV-focused; XR is not discussed.\\
\hline

\cite{10468592} & Quadrotor UAVs & Yes & No & Partial & No & Yes &
Reviews control theory and DT for quadrotor UAV performance. &
Emphasizes control engineering without XR or immersive digital technologies.\\
\hline

\cite{mine} & Mining & Partial & VR/AR & No & No & Yes &
Reviews UAVs, LiDAR, VR/AR, and cloud computing for mine planning. &
Application-specific; lacks comprehensive DT-driven AI framework and general UAV operations.\\
\hline

\textbf{This Survey} & \textbf{Immersive UAVs} & \textbf{Yes} & \textbf{XR} & \textbf{Yes} & \textbf{Yes} & \textbf{Yes} &
\textbf{Comprehensively reviews DT, XR, AI, DRL, Generative AI, UAV modeling, control, simulation-to-reality transfer, swarm coordination, safety, QoS, case studies, and future research directions.} &
\textbf{Provides the first unified survey integrating DT, XR, AI, DRL, GAI, immersive UAV operations, research gaps, and future directions.}\\
\hline

\end{tabularx}
\end{table*}
\subsection{Motivation and Background}

 Digital modeling frameworks such as DT replicate UAV 
 components, processes, dynamics, and 
 behavior in a digital counterpart. The physical and digital counterparts  exchange inputs and operational data through real-time data communications \cite{9899718}. A UAV swarm \cite{10508811} with multiple aerial roles \cite{9314039} can be modeled using DT to enable collaboration, promote safety, and mitigate the simulation-to-reality gap.
\par
XR can significantly improve UAV safety 
 by creating a 3D map of the UAV's surroundings and use such a map to recognize and track objects in the environment \cite{Szstak2024}. In addition, AR can provide navigation and guidance allowing UAVs to navigate complex environments efficiently and safely. However,
XR services impose stringent requirements on 
mobility, data rate, latency, and reliability, increasing the demands on UAVs. Therefore, enabling XR services is a critical UAV challenge.

\begin{figure*} [h]
    \centering 
    \captionsetup{justification=raggedright}
    \includegraphics[scale=0.7]{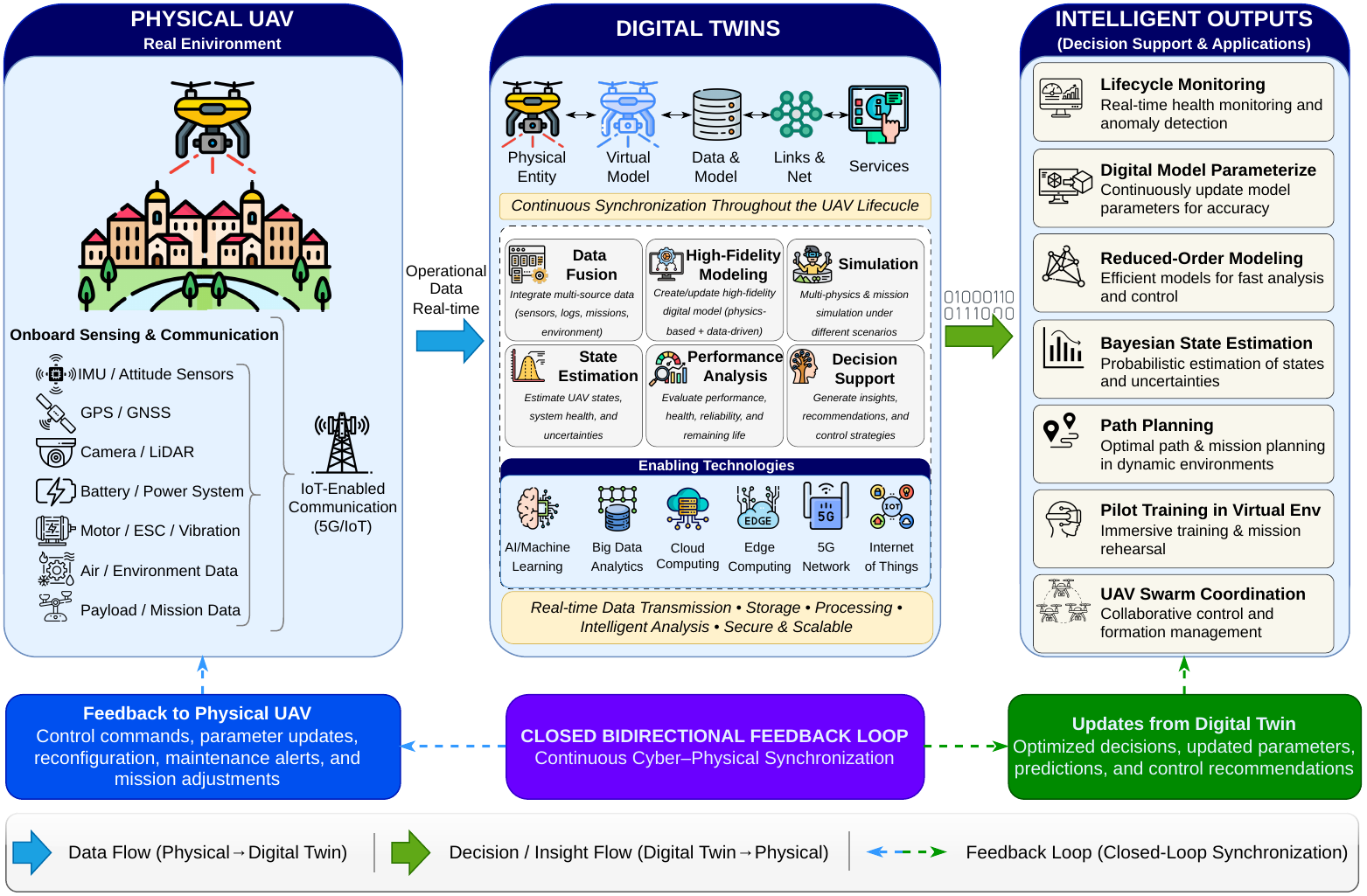}
    \caption{The figure illustrates a Digital Twin framework for an Unmanned Aerial Vehicle (UAV), showing the continuous interaction between the physical UAV and its virtual counterpart through real-time data exchange and intelligent decision support.}
    \label{fig:digital1}
\end{figure*}

\par
The motivation for this work stems from the potential benefits of the integration of DTs and XR in UAVs. This integration enables immersive digital environments that allow UAV operators to interact with replicas of UAV assets in real-time. For example, for a UAV swarm, operators can step into a virtual representation, examine surrounding conditions,
and assess potential changes in swarm parameters without being physically on-site. A salient aspect of merging XR with DT is improved decision-making, where UAV operators can visualize complex data overlays within the XR environment, enhancing situational awareness and facilitating informed action. This is especially useful in environments that are complex or dangerous to access physically, such as harsh and remote areas.
\par
More specifically, we investigate how DTs can be enhanced using Deep Reinforcement Learning (DRL) to develop accurate modeling and control, enable efficient collaboration of UAVs, ensure rigorous safety measures, and bridge the gap between simulations and practice for
effective UAV operations. 

We examine the practical challenges in deployment, such as computational overhead, data synchronization, data management, security, and ethical issues, and highlight potential solutions based on GAI to address some of these challenges. In addition, we explore how XR technologies can improve the operation of UAVs by enhancing environment perception, navigation capabilities, obstacle detection and response to unpredictable scenarios, and facilitating human operator monitoring and intervention. We also discuss practical challenges such as latency, situational awareness, privacy and ethical concerns, as well the role of GAI in improving XR.
\par
Fig. \ref{fig:digital1} shows a DT framework for a UAV, showing the continuous interaction between the physical UAV and its virtual counterpart through real-time data exchange and intelligent decision support.
The Physical UAV operates in the real environment while continuously collecting operational data through onboard sensing and communication systems. The collected information is transmitted through an Internet of Things (IoT)-enabled communication network to maintain synchronization with the digital counterpart. At the center of the figure is the Digital Twin, which consists of five interconnected components: Physical Entity, Virtual Model, Data, Connections, and Services. The virtual model provides a high-fidelity representation of the physical UAV and is continuously updated using real-time operational data together with historical information, enabling accurate monitoring of the UAV throughout its lifecycle. A closed bidirectional feedback loop connects the Digital Twin with the Physical UAV. Operational data continuously update the virtual model, while optimized decisions, updated parameters, and control recommendations are fed back to the physical UAV, forming a continuous cyber–physical synchronization process throughout the UAV lifecycle.
\par
The realization of DT and XR for UAVs rests on a convergence of enabling technologies that have matured in parallel. Low-latency, high-throughput connectivity offered by 5G and emerging 6G networks~\cite{9283775} is perhaps the most critical prerequisite, since meaningful bidirectional synchronization between a physical UAV and its digital counterpart demands near-real-time data exchange. Edge and cloud computing infrastructures complement this by offloading rendering, inference, and model-update tasks away from the UAV's constrained onboard hardware~\cite{10443270}. At the perception layer, advances in computer vision and IoT sensor fusion supply the continuous telemetry stream that keeps DT models calibrated and XR overlays contextually accurate. Underpinning all of these is ML, which transforms raw sensor data into actionable control policies. Together, these technologies define both what is currently achievable and where the next bottlenecks lie.
\par

\begin{figure*} [h]
    \centering 
    \captionsetup{justification=raggedright}
    \includegraphics[width=18cm,height=7cm]{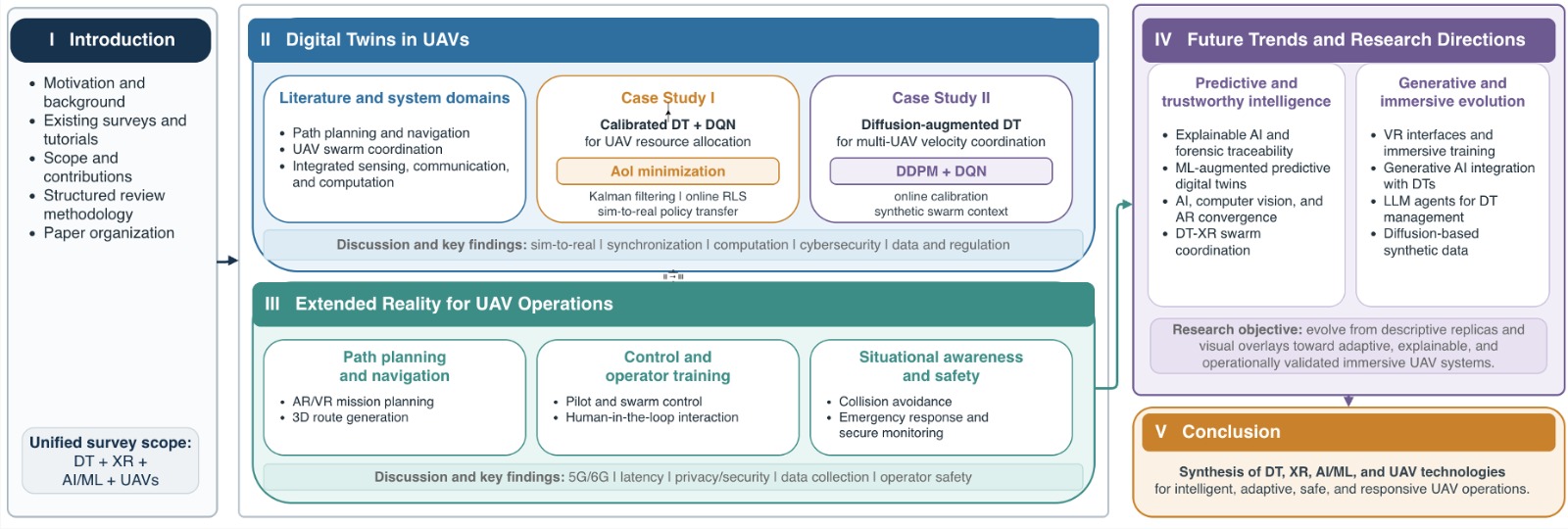}
    \caption{Overall organization and logical flow of the paper, covering the Digital Twin and Extended Reality review streams, the associated case studies and key findings, and the resulting future research directions.}
    \label{fig:digital323}
\end{figure*}

\subsection{Existing Surveys and Tutorials}
When searching the literature for the use of immersive digital technologies in the design, deployment, and operation of UAVs, we found that this topic remains underexplored, particularly in existing surveys. In this section, we briefly review the most closely related surveys and their respective focus. 

Several surveys overlap with our focus but address different targets, particularly within Robotics, e.g., multi-agent systems, and Intelligent Transportation Systems (ITS), e.g., autonomous vehicles. We highlight here those most relevant to our work,
despite their differences. Makhataeva \textit{et al.} \cite{robotics9020021} overviewed AR research in Robotics over five years, categorizing the literature into Medical and Interactive Robotics, Path Planning and Control, and Multi-Agent Systems. Ibrahim \textit{et al.} \cite{su14020601} as well as Ali \textit{et al.} \cite{app13105871}  
reviewed DT technology for ITS, with a particular focus on Electric Vehicles (EVs) and Autonomous Vehicles (AVs). 
Tran \textit{et al.} \cite{9543560} reviewed 31 VR-based studies focused on AV-pedestrian interaction. Kettle \textit{et al.} \cite{kettle_safetyreview} presented a systematic review of AR visualizations for in-vehicle driver communication from 2012 to 2022. 
Riegler \textit{et al.} \cite{riegler2021systematic} provided a review covering 12 years of VR research in the context of automated driving from 2009 to 2020, while Guo \textit{et al.} \cite{GUO2022} reviewed the 
development of the Internet of Vehicles and the use of DTs. 
\par

Turning to surveys that specifically target UAVs, we identify the following. Anandaraj \textit{et al.} \cite{anandaraj2024survey} surveyed the integration of AI and DT technology for UAVs , 
investigating algorithms, learning models, and data processing techniques that improve UAV intelligence and decision-making capabilities. Sarantinoudis \textit{et al.} \cite{10556896} surveyed the specific tools used for DT implementation in the UAV domain. Qin \textit{et al.} \cite{10570372} explored the concepts and potential of DT networks, with a particular focus on the role of ML and particularly DL in improving the efficiency of DTN. Abro \textit{et al.} \cite{10468592} overviewed the contributions of control theory and DTs to the performance of quadrotor UAVs. Piyush \textit{et al.} \cite{mine} reviewed the role of UAV together with LiDAR and VR/AR technologies, supported by cloud computing architectures, in improving mine planning activities. Table \ref{tab:survey_comparison} provides a comparative analysis of existing works.

\subsection{Contribution}
The surveys mentioned in the previous section overlook the role of DTs in UAV modeling and control, in designing collaboration, and in improving safety and Quality of Service (QoS). In contrast, our review addresses these topics, helping to disseminate and advance 
 immersive digital technologies, such as DT and XR, in UAVs.
Modeling, control, and real-world deployment remain challenging, as mismatches between simulations and actual conditions limit real-world effectiveness. We investigate how DT and XR can address these challenges and enhance UAV performance. Moreover, our review highlights the integration of AI algorithms, including DRL and GAI, with these technologies to create more intelligent and adaptive UAV systems. Our review also identifies research gaps and suggests future directions to foster further innovation in this field.

Therefore, we put forward the following contributions:
\begin{itemize}

\item We provide a detailed review of existing literature on the application of DT technology to UAVs, with a specific focus on the integration of ML techniques, particularly DRL. We highlight the role of DTs in enhancing simulation, safety, and operational optimisation for UAVs.
A supporting case study demonstrates that a DT equipped with online Recursive Least-Squares (RLS) channel calibration closes a 28\% path-loss exponent mismatch within 500 training episodes, and the resulting DQN policy transfers to the physical layer without architectural modification. We identify key research gaps and propose countermeasures based on GAI to optimise DT frameworks, further integrate AI, and broaden the range of UAV application scenarios.

  \item We review XR technologies for seamless integration into UAV operations, offering benefits in training, navigation, and situational awareness. We propose a second case study, layering an AR-assisted human-in-the-loop operator intervention protocol atop the DT-DRL pipeline, with a latency-budgeted evaluation protocol for assessing operator response time against system performance. Practical challenges in implementing XR for UAVs such as privacy, safety, and ethical concerns, are discussed. Future research and development efforts are also highlighted, focusing on minimizing latency and risks associated with XR while leveraging  GAI to advance XR capabilities.

\end{itemize}

\subsection{Review Methodology}

This survey was conducted using a structured literature review approach 
drawing primarily on Google Scholar and the IEEE Xplore Digital Library. The search 
used keywords related to UAVs, DT, XR, AI, and DRL. Peer-reviewed journal articles and conference papers were screened based on relevance, technical contribution, and applicability to immersive UAV systems. To ensure coverage of recent advances, priority was given to publications from the last five years while foundational studies were also incorporated where needed for context. The selected literature was then categorized into key themes, namely DT-enabled UAVs, and XR-assisted operations.

\subsection{Paper Structure and Organization}
Section~\ref{sec2} investigates DT technology in the context of UAVs, discusses its integration with DRL, and explores various use cases, 
concluding with several key findings and insights. Section~\ref{sec3} focuses on the role of XR in UAVs, and examines its impact on path planning, collision avoidance, situational awareness, and 
related functionalities. It also concludes with key findings and insights. Section~\ref{sec4} highlights future research directions for research on DT and XR in the context of UAVs. Finally, Section~\ref{sec5} concludes the paper. The overall organization and logical flow of the paper are illustrated in Fig.~\ref{fig:digital323}.

\section{Exploring DTs in UAVs: Simulation, Safety, and Beyond} \label{sec2}
This section reviews the literature on the use of DT in the context of UAVs. It highlights the importance of DT as well as its role in addressing simulation-to-reality problems, and in improving safety.

\begin{table}[h]
\caption{Key Techniques for Path Planning and UAV Navigation}
\label{tab10}
\centering
\scriptsize
\begin{tabular}{|p{0.5cm}|p{1.75cm}|p{1.75cm}|p{3cm}|}
\hline
\textbf{Ref} & \textbf{Technique} & \textbf{Application} & \textbf{Strengths and Limitations} \\
\hline
\cite{SOLIMAN2023106318} & DRL with DT & Path planning and target visiting & Minimizes time and energy overhead; verified on real testbed. \\
\hline
\cite{10.1145/3555661.3560865} & DT-based DRL & Multi-UAV path planning & Facilitates simulation to improve understanding of multi-UAV tasks. \\
\hline
\cite{10076786} & DT-based Visual Navigation & UAV visual navigation & Integrates DT, UAV, and DL; addresses real-world data collection challenges; improves performance and safety. \\
\hline
\cite{9614346} & DRL with DT using BCDDPG & Flocking motion for UAV swarms & Solves practical DRL application problems; DT trains DRL model centrally, enabling rapid deployment in real UAVs. \\
\hline
\cite{10443270} & FRL for DT-empowered UAV-assisted MEC & Task offloading and resource allocation & Secured task offloading and resource allocation using federated reinforcement learning; enhances cooperation among mobile users in dynamic environments. \\
\hline
\end{tabular}
\end{table}
Communication infrastructures are often severely impacted in natural disasters such as earthquakes, wildfires, and hurricanes, leading to 
overload or complete failure in the affected regions. In search and rescue operations, UAVs can function as mobile base stations or relays, offering essential wireless connectivity. The  
deployment and trajectory planning of UAVs are critical to ensuring optimal signal quality and coverage. AI-powered algorithms can improve trajectory optimization by accounting for users' requirements. Their performance 
can be further enhanced by integrating DT technology into the development process, whether to improve UAV design, traffic optimization, QoS or safety~\cite{gurses2024digital}. Once a DT is synchronized with its physical counterpart, it maintains accurate system state information for the learning process. The AI model can the learn from DT and be rapidly deployed in practice. 
High-fidelity simulations enable extensive testing and validation of algorithms under different scenarios and conditions~\cite{10.1145/3555661.3560865,9451579}. DT enables better adaptation to UAV dynamic conditions, such as environmental changes, signal propagation, and user movement by providing a virtual representation of both the UAVs and the surrounding environment. This continuous simulation allows AI algorithms to make more informed adjustments to deployment and trajectory, thereby improving wireless coverage and communication efficiency in disaster-stricken areas~\cite{gurses2024digital}.
\par

\begin{figure}[t]
    \centering
    \includegraphics[width=0.8\linewidth]{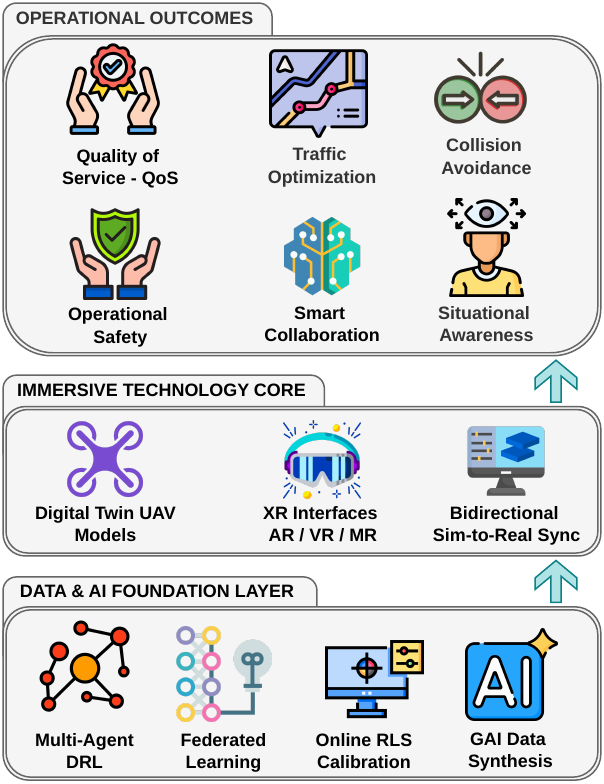}
    \caption{The hierarchical architecture of the immersive UAV management framework, illustrating the integration of the data-driven foundation, the immersive technology core, and the resulting operational outcomes.}
    \label{fig:uav_layer}
\end{figure}

Fig. \ref{fig:uav_layer} illustrates the proposed three-tier architecture for intelligent UAV operations. The Data and AI Foundation Layer provides the essential algorithmic infrastructure, utilizing multi-agent reinforcement learning, federated learning, and generative data synthesis, alongside online RLS calibration to process dynamic environmental inputs. This foundational intelligence directly feeds the Immersive Technology Core, where precise Digital Twin models and XR interfaces are seamlessly coupled through continuous bidirectional sim-to-real synchronization. Ultimately, this integration guarantees the Operational Outcomes at the highest tier, ensuring robust collision avoidance, optimized traffic coordination, Quality of Service (QoS) maintenance, and enhanced situational awareness across the UAV swarm.

\begin{table}[h]
\caption{Key Techniques for Enhancing Cooperative Sensing, Target Search, and Task Assignment in UAV Swarms}
\label{tab20}
\centering
\scriptsize
\begin{tabular}{|p{0.5cm}|p{1.75cm}|p{1.75cm}|p{3cm}|}
\hline
\textbf{Ref} & \textbf{Technique} & \textbf{Application} & \textbf{Strengths and Limitations} \\
\hline
\cite{9263396} & DT with ML & Intelligent cooperation among UAV swarms & Integrated with ML for optimal solution. \\
\hline
\cite{10089851} & DT and DRL-based task assignment & Task assignment for UAV swarms & DRL for reassignment; respects time constraints. \\
\hline
\cite{10045049} & Multi-agent DRL with DT & Cooperative target search by multiple UAVs & Balances training speed and fidelity. \\
\hline
\cite{zhou2024hierarchical} & Hierarchical DT-enhanced cooperative sensing & Cooperative sensing by UAV swarms & Reduces management complexity with edge UAV management; adaptive model aggregation using attention mechanism improves sensing decisions and efficiency. \\
\hline
\cite{cao2025uav} & Multi-agent RL with DT and attention mechanism & Cooperative target search by UAV swarms & Enhances decision-making with attention mechanism; robust sim-to-real deployment validated through physical experiments. \\
\hline
\end{tabular}
\end{table}
\subsection{Literature Review}

\subsubsection{Path Planning and Navigation}

Soliman \textit{et al.}\cite{SOLIMAN2023106318} formulated path planning and 
target visiting as an optimization problem and proposed a DRL strategy to learn target patterns while minimizing UAV time and energy overhead. They developed a Gazebo simulator-based DT to overcome the constraints of the RL exploration phase and verified the feasibility of 
the framework by developing a real testbed. Building on this idea, Li et al. \cite{10.1145/3555661.3560865} investigated UAV path planning using DT technology to address limitations in DRL models. Traditional simulations often fail to capture real-world complexity, reducing the effectiveness of DRL-based models. By integrating DT with DRL, a virtual replica of the UAV and its environment enables realistic simulations that provide real-time data and feedback from actual operations. This enhances DRL training, allowing models to adapt to factors 
such as terrain, obstacles, wind, and other dynamic conditions.

The result is more reliable and efficient trajectory planning for UAVs, 
enabling safer navigation in practical scenarios.
\par

Similarly, Miao et al. \cite{10076786} proposed a UAV visual navigation system that integrates DT, UAV, and DL to address real-world data collection challenges and improve UAV monitoring, performance, and safety. Their DT framework comprises four layers: (1) Physical Space, where a visual navigation strategy runs on the UAV’s onboard computer; (2) Digital Space, which includes accurate 3D models 
and virtual UAVs; (3) Twin Data Layer, which bridges physical and digital spaces by facilitating data collection, monitoring, control, and fault reproduction, using a MySQL database for real-time processing; and (4) Application Layer, which integrates models, data, and algorithms into user-centric tools and interfaces, enabling intelligent navigation, interaction, and multi-platform hardware-in-the-loop functionalities.

Moreover, Shen \textit{et al.}\cite{9614346} studied flocking motion for UAV swarms 
using DRL. A DT built on the DRL training framework 
leveraged to address the practical problems of DRL applications. The DT was incorporated into a central system to train the DRL model, which was then deployed on the physical and digital units. The DRL algorithm is  an actor-critic, namely the Behavior-Coupling Deep Deterministic Policy Gradient (BCDDPG). 
Finally,  Consul \textit{et al.} \cite{10443270} addressed the challenge of task offloading and resource allocation in a DT-empowered UAV-assisted MEC network with user movement using the adaptive DT framework to predict the dynamic network environment accurately. DT assists in improving mobile users' cooperation.
Federated Reinforcement Learning (FRL) was employed to secure task offloading and optimize resource allocation for the DT-empowered UAV-assisted MEC system. In this system, multiple UAVs equipped with MEC servers and multiple resource devices support the FRL process, while mobile users randomly generate computing tasks as they move. Table \ref{tab10} summarizes the techniques discussed above.  

Collectively, these studies demonstrate the value of integrating DT frameworks with DRL for UAV path planning. Frameworks using centralized DT training architectures \cite{9614346} improve policy convergence and transferability but incur higher computational load and synchronization latency. In contrast, federated schemes such as \cite{10443270} enhance privacy and cooperation among UAVs but face scalability challenges due to communication overhead. 
The DT also contributes to 6G UAV communications by providing a scalable and modular framework for evaluating performance, generating valuable data, and laying the groundwork for future intelligent, autonomous network optimization that is critical for the ultra-reliable, low-latency requirements of 6G\cite{11152769}.

\subsubsection{UAV Swarm}

Shen  \textit{et al.} \cite{10045049} proposed a multi-agent DRL approach for UAV swarms to cooperatively search for static targets in dynamic threat environments, despite limited sensing and communication capabilities. They introduced a centralized DT training framework with decentralized execution to balance training speed and environmental fidelity. The DT system consists of: (1) Physical Entity, representing real-world UAVs; (2) Digital Model, a virtual counterpart of the UAVs; (3) Decision Model, which governs decision-making using insights from the digital model; and (4) Connections, enabling communication and data exchange between physical and digital components. During training, multiple DT environments collect data simultaneously for centralized 
learning, improving efficiency. Once trained, the models are deployed on real UAVs for distributed target search.

\par
In parallel, Lei \textit{et al.}\cite{9263396} explored the use of 
DT technology to enhance UAV swarm collaboration by providing a high-fidelity virtual model for real-time monitoring and optimization. This DT model enables simulation of collaborative scenarios in a controlled virtual environment, improving real-world behavior while minimizing resource-intensive computations on UAVs. Integrating ML algorithms into the DT framework 
enables exploration of 
near-optimal solutions for cooperative decision-making, which can then guide swarm behavior. The DT model overcomes UAV computational and storage constraints by offloading heavy processing tasks to a central system, enabling complex tasks  
such as intelligent network reconstruction. This approach improves coordination, adaptability, and efficiency
for UAV swarm operations in dynamic environments.

\par
Moreover, Tang \textit{et al.}\cite{10089851} argued that the temporal constraints of a UAV swarm can lead to a low-task completion ratio. In this context, they proposed a DT and DRL-based task assignment method. In the initial task assignment, 
an airship uses a genetic algorithm to divide a task into multiple subtasks based on to the shortest distance and assign them to UAVs. In task reassignment,
a DT model is established in the airship to address the Size, Weight and Power (SWAP)  
constraints of small-sized UAVs.  
A DQN with 
dynamic Markov properties the determines
the UAV's behavior to reassign the task, improving 
task completion while respecting the time constraint. Performance evaluation confirmed that the training period of the DQN with DT 
was shortened. Additionally, Zhou \textit{et al.} \cite{zhou2024hierarchical} proposed a hierarchical DT-enhanced cooperative sensing architecture and an adaptive model aggregation algorithm to improve UAV-based collaborative perception. The architecture reduces management complexity by allowing edge UAVs to manage local UAV groups and aggregate lightweight DT models built 
from local physical information at low computational cost. To ensure low-latency detection, the UAVs predict the trajectories of physical objects and coordinate position and attitude adjustments in advance. Edge UAVs also enable the real-time exchange of parameters between regions to improve detection accuracy. The adaptive model aggregation algorithm uses an attention mechanism to select highly correlated DT models and instruct the UAVs to dynamically adjust their positions, attitudes, and associations with physical objects. This approach optimizes sensing decisions yielding efficient sensing paths and improving overall cooperative perception.

\par

Meanwhile, Gomes \textit{et al.} \cite{gomes2026integrating} 
presented
a DT-based UAV surveillance framework that integrates swarm intelligence and computer vision for wildfire monitoring. By combining Boids-inspired patrol path generation with autonomous and manual flight modes, the system improves area coverage, reduces blind spots, and enhances detection reliability. Automated entity recognition, evidence storage, and a web interface further support practical mission management and post-mission analysis.
The proposed architecture addresses a key gap in existing research by unifying UAV swarms, DTs, and intelligent patrol strategies within a single extensible system for wildfire prevention and detection.

\begin{table}[htbp]
\centering
\caption{Taxonomy of Sim-to-Real Transfer Mechanisms in UAV Digital Twins}
\label{tab:sim2real_taxonomy}
\begin{tabular}{|p{0.5cm}|p{1.5cm}|p{5.7cm}|}
\hline
\textbf{Ref} & \textbf{Category} & \textbf{Description} \\
\hline
\cite{SOLIMAN2023106318}, \cite{9614346} & Direct policy transfer & A DRL policy trained entirely in a DT or simulator is deployed on the physical UAV with no further adaptation. \\
\hline
\cite{10045049}& Centralized training + decentralized execution & A central DT trains multi-agent policies; each real UAV runs the same policy locally without central runtime control. \\
\hline
\cite{cao2025uav} & Twin training + continuous evolution & A DT-driven framework where training, deployment, and ongoing evolution are coupled (“twin training, decentralized execution, continuous evolution”). \\
\hline
\cite{gurses2404digital}  & Calibrated DT development & Real-world data is used to calibrate the DT \textit{before} training or deployment; the calibrated DT then generates more realistic training conditions. \\
\hline
\cite{cao2025uav} & Bidirectional real-time synchronization &
  The DT is continuously updated with real-time sensor data from the physical UAV,
  enabling state estimation and monitoring, but policy adaptation is not explicitly
  described. \\
\hline
This work &
  Bidirectional sync with online parametric calibration &
  Noisy telemetry (GPS, path-loss) is Kalman-filtered at every synchronisation
  event and fed into an online RLS estimator that updates the DT's path-loss
  exponent in real time. The trained DQN policy transfers to the physical UAV
  without modification, achieving sim-to-real transfer through calibration
  rather than domain adaptation or fine-tuning. \\
\hline
\end{tabular}

\end{table}

Finally, Cao \textit{et al.}\cite{cao2025uav} introduced a scalable cooperative target search algorithm that leverages multi-agent RL and a multi-head attention mechanism to enhance decision-making efficiency. Each agent utilizes a graph-based observation space and an environmental cognition map to optimize target search rate, area coverage, and safety. A DT-driven training framework, "twin training, decentralized execution, and continuous evolution," addresses the sim-to-real gap by deploying multi-agent DRL-trained policies to real-world UAV swarms for coordinated signal source search. Additionally, a DT-enabled validation system integrates real UAV flight control, communication network simulations, and a 3D physics engine, with physical experiments demonstrating the system's scalability and robustness across varying  
numbers of UAVs and mission area sizes. Table~\ref{tab20} summarizes the above findings. Table \ref{tab:DT_DRL_comparison} presents a comparative summary of the most representative studies that integrate DT technology with DRL for UAV applications. Collectively, these works illustrate how DT environments are being used to simulate flight conditions, generate training data, and accelerate policy learning for tasks such as path planning, swarm coordination, and resource optimization. As reflected in the table, the field has matured in DT model design and DRL algorithm selection, but it still requires deeper quantitative evaluation—particularly regarding computational cost, latency, and simulation-to-reality performance, to enable reliable deployment of DT-empowered UAV systems.
\par
Multi-agent DRL frameworks, such as those in \cite{10045049} and \cite{cao2025uav}, use centralized training and decentralized execution architectures, enabling improved decision-making and target search efficiency. However, they introduce significant computational complexity and require centralized DT environments, which may not 
scale efficiently to real-world distributed systems. Hierarchical and attention-based models \cite{zhou2024hierarchical} enhance sensing accuracy and resource management. Genetic and DQN-based hybrid methods \cite{10089851} achieve adaptive reassignment but struggle with convergence stability under dynamic task loads. Table \ref{tab:sim2real_taxonomy} provides a taxonomy of sim-to-real transfer mechanisms.

\begin{table*}[htbp]
\centering
\caption{Comparative analysis of DT + DRL frameworks for UAVs}
\label{tab:DT_DRL_comparison}
\footnotesize
\begin{tabularx}{\textwidth}{@{}l>{\raggedright\arraybackslash}X>{\raggedright\arraybackslash}Xl>{\raggedright\arraybackslash}X>{\raggedright\arraybackslash}X@{}}
\toprule
\textbf{Ref.} & \textbf{Problem} & \textbf{Simulator/Dataset} & \textbf{Validation} & \textbf{Key strengths} & \textbf{Limitations} \\
\midrule
\cite{SOLIMAN2023106318} & Path planning and target visiting using DRL & Gazebo-based DT & Sim + Testbed & Realistic DT; hardware verified & Runtime/sample-efficiency not reported \\
\midrule
\cite{10.1145/3555661.3560865} & Multi-UAV path planning via DT-assisted DRL & Custom DT simulator & Simulation only & Reduces sim-to-real gap & No field validation \\
\midrule
\cite{10076786} & Visual navigation with DT-generated imagery & DT image dataset & Simulation/Dataset & Safe data generation; improved performance & No latency/model-size data \\
\midrule
\cite{9614346} & Swarm coordination and flocking control & DT environment & Simulation only & Centralized DT training & Compute-heavy; latency unreported \\
\midrule
\cite{10443270} & Task off-loading and resource allocation & Network/MEC simulator & Simulation only & Privacy-preserving DT + FL & FL overhead not quantified \\
\midrule
This work & Joint sensor scheduling and velocity control; AoI
  minimisation with online DT calibration &
  Custom DT with Kalman filtering, RLS calibration, and
  z-score anomaly detection; PhysicalUAVStub with
  deliberate 28\% channel mismatch &
  Sim + PhysicalUAVStub (Phase~3) &
  Online RLS closes path-loss exponent gap; anomaly
  detection; clean sim-to-real handover without policy
  modification &
  Single UAV; fixed circular trajectory; stub not
  connected to real hardware \\
\bottomrule
\end{tabularx}
\end{table*}

\subsection{Integrated sensing, communication, and computation networks}

Li \textit{et al.}~\cite{10190734}  studied the use of DTs to improve the performance of Integrated Sensing, Communication, and Computation (ISCC) networks. Users conduct radar sensing and computational task offloading over the same frequency band, while UAVs are deployed as mobile edge computing servers.  
Radar sensing accuracy and the energy consumption of computation offloading are jointly optimized.
The multi-objective optimization is formulated as follows: 

\textbf{Objective Functions:}

1. Minimize the weighted energy consumption:
\begin{equation}
\min_{W_c, W_r, q, \tilde{f}_i, \tilde{f}_e, \rho} \;
  \omega \sum_{n=1}^N \sum_{m=1}^M E_m[n]
  + \sum_{n=1}^N \sum_{k=1}^K E_k[n].
\label{eq:iscc_obj_energy}
\end{equation}
where:
\begin{itemize}
    \item $N$, $M$, and $K$ denote the numbers of time slots, UAVs, and users, respectively. For convenience, let $\mathcal{N}=\{1,\dots,N\}$, $\mathcal{M}=\{1,\dots,M\}$, and $\mathcal{K}=\{1,\dots,K\}$ denote the corresponding index sets.
    \item $W_c$: Precoding matrix of communication symbols for all users and time slots.
    \item $W_r$: Precoding matrix of radar waveforms for all users and time slots.
    \item $q$: Trajectory planning variables for all UAVs and time slots.
    \item $\tilde{f}_i$: CPU frequency variables for all users and time slots.
    \item $\tilde{f}_e$: CPU frequency allocation variables for UAVs to all users and time slots.
    \item $\rho$: Task partition factors for all users and time slots.
    \item $E_m[n]$: Energy consumption of UAV $m$ at time slot $n$.
    \item $E_k[n]$: Energy consumption of user $k$ at time slot $n$.
    \item $\omega$: non-negative weight factor for the UAV energy consumption (relative to users).
\end{itemize}

2. Minimize the beampattern performance deviation:
\begin{equation}
\min_{W_c, W_r} \sum_{n=1}^N \sum_{k=1}^K \| X_k[n] - R_{d,k} \|_F^2.
\label{eq:iscc_obj_beam}
\end{equation}
where:
\begin{itemize}
    \item $X_k[n]$: Transmission covariance matrix of user $k$ at time slot $n$.
    \item $R_{d,k}$: Desired radar beampattern covariance matrix for user $k$.
    \item where $\|\cdot\|_F$ denotes the Frobenius norm.
\end{itemize}

\textbf{Constraints:}

1. \textbf{User Association}:
\begin{equation}
\alpha_{k,m} \in \{0, 1\}, \quad
\sum_{m=1}^M \alpha_{k,m} \leq 1, \quad
\forall k \in \mathcal{K}, \; \forall m \in \mathcal{M}.
\label{eq:iscc_c_assoc}
\end{equation}

Each user $k$ can associate with at most one UAV $m$. The binary variable $\alpha_{k,m}$ equals 1 if user $k$ is associated with UAV $m$, and 0 otherwise.

2. \textbf{Transmission Power}:
\begin{equation}
\mathrm{tr}\!\left(X_k[n]\right) \leq p_{\max}, \quad
\forall k \in \mathcal{K}, \; \forall n \in \mathcal{N}.
\label{eq:iscc_c_power}
\end{equation}

The total transmission power of user $k$ at time slot $n$ cannot exceed the maximum allowable power $p_{\text{max}}$.

3. \textbf{Interference-to-Noise Ratio (INR)}:
\begin{equation}
\eta_k[n] \leq \zeta_k, \quad
\forall k \in \mathcal{K}, \; \forall m \in \mathcal{M}.
\label{eq:iscc_c_inr}
\end{equation}

The interference-to-noise ratio (INR) for each user $k$ at time slot $n$ must be below the maximum tolerable level $\zeta_k$.

4. \textbf{Computation Delay}:
\begin{equation}
\max\!\left\{ t_k^l[n],\; t_k^o[n] + t_k^e[n] \right\}
  \leq t_k^{\max}[n], \quad
\forall k \in \mathcal{K}, \; \forall n \in \mathcal{N}.
\label{eq:iscc_c_delay}
\end{equation}

The total computation delay, including local ($t_k^l[n]$) and offloading delay ($t_k^o[n] + t_k^e[n]$), must not exceed the maximum allowable delay $t_k^{\text{max}}[n]$.

5. \textbf{CPU Frequency of Users}:
\begin{equation}
0 \leq \tilde{f}_k^l[n] \leq f_k^{\max}, \quad
\forall k \in \mathcal{K}, \; \forall n \in \mathcal{N}.
\label{eq:iscc_c_cpu_user}
\end{equation}

The CPU frequency of each user $k$ is bounded by the maximum available frequency $f_k^{\text{max}}$.

6. \textbf{CPU Frequency Allocation of UAVs}:
\begin{subequations}\label{eq:iscc_c_cpu_uav}
\begin{align}
  0 &\leq \tilde{f}_{k,m}[n] \leq f_m^{\max},
    \quad \forall k \!\in\! \mathcal{K},\, \forall m \!\in\! \mathcal{M},\, \forall n \!\in\! \mathcal{N},
    \label{eq:iscc_c_cpu_uav_a}\\[4pt]
  0 &\leq \sum_{k=1}^K \alpha_{k,m}\,\tilde{f}_{k,m}[n] \leq f_m^{\max},
    \quad \forall m \!\in\! \mathcal{M},\, \forall n \!\in\! \mathcal{N}.
    \label{eq:iscc_c_cpu_uav_b}
\end{align}
\end{subequations}

The allocated CPU frequency $\tilde{f}_{k,m}[n]$ for user $k$ by UAV $m$ must not exceed the UAV's total computational capacity $f_m^{\text{max}}$.

7. \textbf{Task Partitioning}:
\begin{equation}
0 \leq \rho_k[n] \leq 1, \quad
\forall k \in \mathcal{K}, \; \forall n \in \mathcal{N}.
\label{eq:iscc_c_partition}
\end{equation}

The task partition factor $\rho_k[n]$ represents the fraction of the task offloaded by user $k$ and is bounded between 0 and 1.

8. \textbf{UAV Acceleration and Speed Limits}:
\begin{equation}
\| a_m[n] \| \leq a_{\max}, \quad
\| v_m[n] \| \leq v_{\max}, \quad
\forall m \in \mathcal{M}, \; \forall n \in \mathcal{N}.
\label{eq:iscc_c_speed}
\end{equation}

The acceleration and velocity of UAV $m$ are constrained by their respective maximum limits, $a_{\text{max}}$ and $v_{\text{max}}$.

9. \textbf{Minimum Safe Distance Between UAVs}:
\begin{equation}
\| q_i[n] - q_j[n] \|^2 \geq d_{\min}^2, \quad
\forall i, j \in \mathcal{M}, \; i \neq j.
\label{eq:iscc_c_dist}
\end{equation}

The minimum distance $d_{\text{min}}$ between UAVs $i$ and $j$ must be maintained to avoid collisions.

The predictive capabilities of the DT are utilized, even in the presence of estimation errors. The joint optimization
problem~\eqref{eq:iscc_obj_energy}--\eqref{eq:iscc_c_dist} is formulated as a multi-agent Markov decision process (MDP) and addressed using the Multi-Agent Proximal Policy Optimization (MAPPO) method. To improve training performance, techniques such as Beta-distribution policies \cite{chou2017beta} and attention mechanisms \cite{vaswani2017attention} are incorporated, allowing faster convergence and enhanced outcomes.  
\par
In ISCC networks, users first conduct radar sensing to gather multi-view data, which is then uploaded to edge computing servers to support low-latency services. In distributed DT-enabled ISCC networks, a framework combining centralized training with decentralized execution is employed. In this framework, agents take actions using their actor networks while the centralized critic networks are trained for each agent type.
Since rewards are often dependent on the actions and states of multiple agents, which are difficult for users and UAVs to evaluate independently, the DT plays a key role. During operation, observations and actions are collected from the agents as they interact with the physical environment. These data are then sent to the DT layer to update the virtual twins. Using this information, the DT layer evaluates rewards by calculating metrics like actual computing time and energy consumption. Additionally, the DT layer dynamically estimates key hyperparameters, such as the task information distribution, in real time. The DT layer combines all observations into a global state and sends it to the critic networks for each type of agent. These critic networks then act as centralized state-value functions to guide decision-making effectively.  
\par
Numerical results demonstrate that the proposed approach effectively balances radar beampattern performance~\eqref{eq:iscc_obj_beam} and computational energy consumption~\eqref{eq:iscc_obj_energy}. Additionally, energy efficiency with respect to~\eqref{eq:iscc_obj_energy} is improved compared to existing methods, emphasizing the potential of DTs in managing complex networks.

\par

The integration of DT and ML in UAVs is increasingly driven by the need to improve UAV safety, navigation, and operational efficiency. DTs bridge the gap between simulated and real environments, enabling precise control and advanced decision-making processes in dynamic UAV scenarios. However, as the need for immersive and real-time interaction with UAV systems continues to grow, the role of XR technologies becomes increasingly significant. XR offers an immersive interface for enhancing safety, navigation, and operator training thus complementing the advancements achieved through DTs. The following section explores how XR technologies revolutionize UAV functionalities, particularly in path planning~\cite{9058989}, navigation~\cite{9847217}, and collision avoidance.

\subsection{Case Study I: DT-Driven DRL for UAV Resource Allocation}
\label{sec:casestudy}

\begin{figure*}[h]
    \centering
    \includegraphics[width=0.9\textwidth]{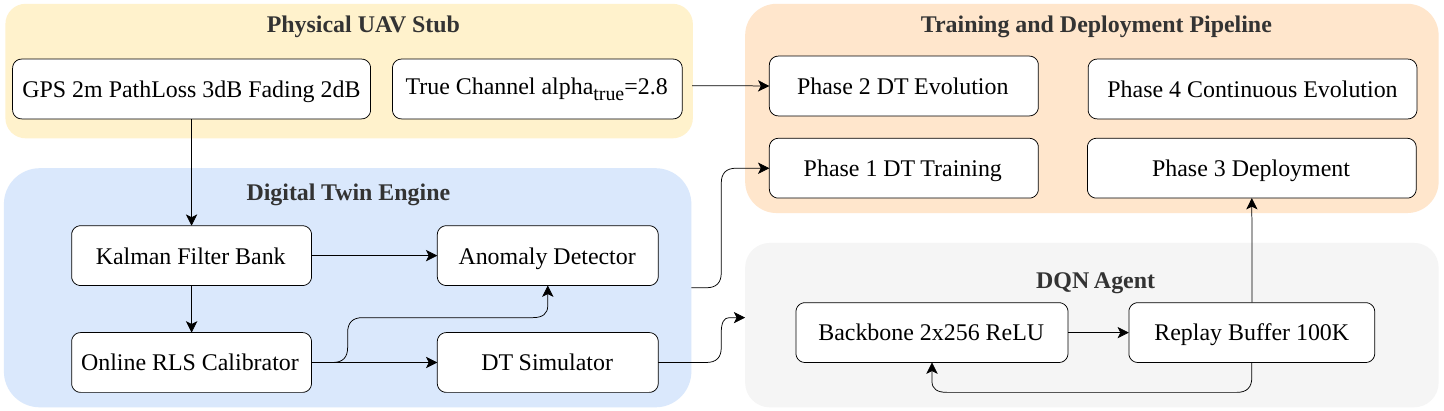}
    \caption{Overview of the DT-driven DRL pipeline for UAV resource allocation, illustrating the Physical UAV Stub, Digital Twin Engine with Kalman filtering, online RLS calibration, and anomaly detection, the DQN Agent with dual action heads, and the four-phase training and deployment pipeline.}
    \label{fig:casestudy_overview}
\end{figure*}

This case study presents a calibrated DT-driven DRL pipeline for joint UAV velocity control and sensor scheduling in a UAV-assisted IoT network. Unlike static simulators, the proposed DT architecture dynamically calibrates its internal models; the DT's channel model is deliberately inaccurate at initialization and continuously adapts using noisy emulated telemetry data during training. One UAV collects data from $N\!=\!10$ ground sensors distributed uniformly within a circular area of radius $r\!=\!400$\,m. The UAV follows a discretised circular trajectory at altitude $h\!=\!100$\,m, and at each step simultaneously selects which sensor to serve and the flight speed, with the objective of minimizing the mean Age of Information (AoI) across all sensors, as illustrated  in Fig.~\ref{fig:casestudy_overview}.

\medskip
\noindent\textbf{Physical Layer and the Sim-to-Real Gap.}
Three independent noise processes are present in the physical layer:
GPS position error
  $\delta_{\mathrm{pos}} \sim \mathcal{N}(0,\,\sigma_p^2)$, $\sigma_p\!=\!2$\,m;
path-loss measurement noise
  $\delta_m \sim \mathcal{N}(0,\,\sigma_m^2)$, $\sigma_m\!=\!3$\,dB; and
random channel fading
  $\delta_f \sim \mathcal{N}(0,\,\sigma_f^2)$, $\sigma_f\!=\!2$\,dB.
The critical mismatch is in the path-loss exponent:
the physical channel has $\alpha_{\mathrm{true}}\!=\!2.8$,
while the DT is initialised with $\hat{\alpha}_0\!=\!2.0$~---~a 28\% error that the calibration mechanism must resolve. Both layers share the path-loss model
\begin{equation}
  \mathrm{PL}(d) \;=\;
    \underbrace{20\log_{10}\!\!\left(\frac{4\pi d}{\lambda}\right)}_{\mathrm{PL_{FS}}(d)}
    + \hat{\alpha}\cdot 10\log_{10}(d),
  \label{eq:pl_model}
\end{equation}
where $d$ is the 3-D UAV-to-sensor distance and $\lambda\!=\!c/f_c$ at $f_c\!=\!2$\,GHz; the physical layer evaluates~\eqref{eq:pl_model} with $\alpha_{\mathrm{true}}$, while the DT uses its running estimate $\hat{\alpha}$, the excess exponent beyond free-space propagation.

\medskip
\noindent\textbf{State Estimation via Kalman Filtering.} Because both position and channel readings are corrupted by noise, the DT maintains
a bank of independent scalar Kalman filters: two for UAV coordinates $(x,\,y)$ and one per sensor for path-loss.
For each tracked scalar $x$, the predict--update cycle is
\begin{align}
  P_{k|k-1}  &= P_{k-1|k-1} + Q,                              \label{eq:kf_pred}  \\
  K_k         &= \frac{P_{k|k-1}}{P_{k|k-1} + R},             \label{eq:kf_gain}  \\
  \hat{x}_k   &= \hat{x}_{k-1} + K_k\!\left(z_k-\hat{x}_{k-1}\right),
    \label{eq:kf_state} \\
  P_{k|k}     &= \left(1 - K_k\right)P_{k|k-1},               \label{eq:kf_cov}
\end{align}
where $z_k$ is the noisy measurement, $P$ the error covariance, $Q$ the process noise, and $R$ the measurement noise. Position filters are configured with $Q\!=\!1$\,m$^2$, $R\!=\!\sigma_p^2\!=\!4$\,m$^2$; path-loss filters with $Q\!=\!0.5$\,dB$^2$, $R\!=\!\sigma_m^2\!=\!9$\,dB$^2$.

\medskip
\noindent\textbf{Online Calibration via Recursive Least Squares.}
At each synchronisation event the DT feeds the Kalman-fused measurements into an online RLS estimator for $\hat{\alpha}$.
Rearranging~\eqref{eq:pl_model} reveals that the unknown exponent enters linearly:
\begin{equation}
  \underbrace{\mathrm{PL}_{\mathrm{meas}} - \mathrm{PL_{FS}}(d)}_{\bar{y}}
  \;=\;
  \underbrace{10\log_{10}(d)}_{\phi}\,\hat{\alpha} \;+\; e,
  \label{eq:rls_model}
\end{equation}
yielding the scalar regression $\bar{y} = \phi\,\hat{\alpha} + e$. With forgetting factor $\lambda_f\!=\!0.98$ the RLS update is
\begin{align}
  K^{\mathrm{RLS}}  &= \frac{P^{\mathrm{RLS}}\,\phi}
                           {\lambda_f + \phi^2 P^{\mathrm{RLS}}},
                    \label{eq:rls_gain}  \\
  \hat{\alpha}      &\leftarrow \hat{\alpha}
                      + K^{\mathrm{RLS}}\!\left(\bar{y} - \phi\,\hat{\alpha}\right),
                    \label{eq:rls_alpha} \\
  P^{\mathrm{RLS}}  &\leftarrow \frac{1}{\lambda_f}
                      \!\left(1 - K^{\mathrm{RLS}}\phi\right)P^{\mathrm{RLS}},
                    \label{eq:rls_cov}
\end{align}
with $\hat{\alpha}$ clipped to $[1.5,\,5.0]$ after each update to enforce physical plausibility. The forgetting factor discounts older measurements, allowing the DT to track slow environmental drift even after initial convergence.

\medskip
\noindent\textbf{Anomaly Detection.}
The DT continuously compares its predicted path-loss against the physically received value. Let $r_k = \mathrm{PL}^{\mathrm{meas}}_k - \mathrm{PL}^{\mathrm{DT}}_k$ be the residual at synchronisation step $k$, tracked over a sliding window of $W\!=\!10$ samples. An anomaly flag is raised when the standardised residual exceeds the three-sigma threshold:
\begin{equation}
  z_k \;=\; \frac{\left|r_k - \bar{r}\right|}{\sigma_r} \;>\; \gamma, \quad \gamma = 3,
  \label{eq:anomaly}
\end{equation}
where $\bar{r}$ and $\sigma_r$ are the window mean and standard deviation. A condition satisfying~\eqref{eq:anomaly} is statistically inconsistent with the calibrated noise floor and points to events such as GPS spoofing, False Data Injection (FDI) on sensor telemetry, or hardware faults.

\medskip
\noindent\textbf{DQN Architecture.}
A shared two-layer backbone (256 units per layer, ReLU activations) feeds two output heads: one over the $N$ sensor choices and one over the 15 discrete velocity levels. The state vector is defined in Table~\ref{tab:state_merged}; the complete
set of hyperparameters (DQN training, Kalman filtering, RLS calibration,
and anomaly detection) is provided in Table~\ref{tab:full_params} in the
Appendix.

\begin{table}[h]
\centering
\caption{Definition of the 13-dimensional state space.}
\begin{tabular}{|c|c|p{5cm}|}
\hline
\textbf{State} & \textbf{Dimension} & \textbf{Meaning} \\
\hline
\multirow{2}{*}{$s[0]$ to $s[9]$} & \multirow{2}{*}{10} &
  Current AoI value for each of the 10 sensor nodes \\
\hline
$s[10]$ & 1 & Path loss in dB to the most recently selected node \\
\hline
$s[11]$ & 1 & UAV $x$-coordinate at the current waypoint (metres) \\
\hline
$s[12]$ & 1 & UAV $y$-coordinate at the current waypoint (metres) \\
\hline
\end{tabular}
\label{tab:state_merged}
\end{table}

\begin{figure*}[ht]
  \centering
  \includegraphics[width=\textwidth]{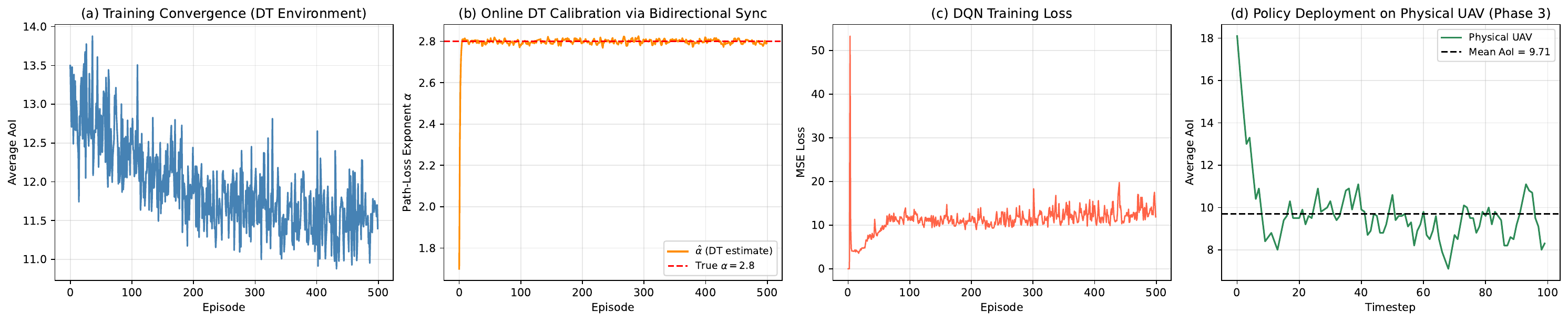}
  \caption{Results of the four-phase DT-driven DRL pipeline for UAV resource allocation.
    \textbf{(a)}~Training convergence: mean AoI per episode inside the DT, declining from approximately 12.5 in the exploration-dominated early episodes toward ${\approx}\,11.5$ as the policy matures.
    \textbf{(b)}~Online calibration: the DT's path-loss exponent estimate
    $\hat{\alpha}$ converges from its erroneous initial value of 2.0 to 2.792,
    reaching within 0.3\% of the true physical value
    $\alpha_{\mathrm{true}}\!=\!2.8$ through 3\,000 bidirectional
    synchronisation events~---~direct evidence that a genuine DT, not a static
    simulator, is in operation.
    \textbf{(c)}~DQN training loss (Bellman MSE), stabilising as the policy converges.
    \textbf{(d)}~Deployment performance on the physical UAV stub (Phase~3):
    with exploration disabled, the policy achieves a mean AoI of 9.26, lower than the training-phase average and confirming successful generalisation across the physical-virtual channel gap.
  }
  \label{fig:dt_results}
\end{figure*}


\noindent\textbf{Four-Phase Training and Deployment Pipeline.}
Training proceeds in four phases, with the first two running concurrently:
\begin{enumerate}
  \item \textbf{DT Training (Phase~1):} The DQN agent interacts exclusively with the
    DT, building experience in a replay buffer and updating its weights via
    a standard soft-update target network scheme.

  \item \textbf{DT Evolution (Phase~2):} In parallel, the physical UAV stub provides
    telemetry every $\Delta_{\mathrm{sync}}\!=\!5$ decision steps.
    Each incoming packet is first filtered through
    Equations~\eqref{eq:kf_pred}--\eqref{eq:kf_cov}
    and then fed into the RLS calibrator
    (Equations~\eqref{eq:rls_gain}--\eqref{eq:rls_cov}),
    continuously pulling $\hat{\alpha}$ toward $\alpha_{\mathrm{true}}$.
    Over 500 training episodes this generates 3\,000 synchronisation events.
  \item \textbf{Deployment (Phase~3):} The trained policy is transferred to the
    physical UAV stub without any architectural changes, demonstrating a clean
    sim-to-real handover enabled by the DT's now-calibrated channel model.
  \item \textbf{Continuous Evolution (Phase~4):} Synchronisation continues during
    deployment, keeping the DT aligned with the physical channel and allowing the
    system to adapt to future environmental changes.
\end{enumerate}


Fig.~\ref{fig:dt_results} summarises the pipeline across all four phases. Panel~(a) shows that the DQN policy progressively reduces mean AoI, with the steep early decline marking the transition from random exploration to policy exploitation. Panel~(b) provides the clearest evidence that the system goes beyond plain simulation: $\hat{\alpha}$ rises monotonically from 2.0 to 2.792, converging to within 0.3\% of $\alpha_{\mathrm{true}}\!=\!2.8$. This calibration happens \emph{passively} alongside DQN training as a direct consequence of the periodic synchronisation events, requiring no separate calibration phase or additional interaction budget. Panel~(c) shows the expected reduction in Bellman error as the target network stabilises the training signal. Panel~(d) records the deployment outcome: mean AoI drops from ${\approx}\,11.5$ (training) to 9.26 on the physical UAV stub,
confirming that the policy generalizes across the sim-to-real channel gap introduced by the 28\% initial exponent mismatch. The anomaly detector~\eqref{eq:anomaly} produced zero false positives across all 3\,000 synchronisation events under normal channel conditions,
validating that $\gamma\!=\!3$ is an appropriate threshold for the noise
levels used in this experiment.

\subsection{Case Study II: Diffusion-Augmented Digital Twin for Multi-UAV Velocity Coordination}
\label{sec:casestudy_diff}
 
Whereas the DT of Section~\ref{sec:casestudy} replays calibrated dynamics directly, here it hosts a conditional \emph{generative} model that synthesises the swarm context used for policy training. The task also differs: velocity-space separation within a small swarm rather than AoI
minimisation, but the calibration machinery (Kalman-filtered synchronisation and online RLS) carries over unchanged.
 
\medskip
\noindent\textbf{Setup.}
One ego UAV coordinates with $J\!=\!3$ neighbours, each described by a velocity vector $\mathbf{v}^{(j)}_t \in [1,15]^4$ (m/s), while the ego UAV applies a common speed command $v^{\mathrm{own}}_t \in \{1,\dots,15\}$. Each neighbour-velocity component follows mean-reverting AR(1) dynamics,
\begin{equation}
  v^{(j)}_{t+1} = \theta_{\mathrm{true}}\, v^{(j)}_{t}
  + \bigl(1-\theta_{\mathrm{true}}\bigr)\mu_{\mathrm{true}} + w_t,
  \qquad w_t \sim \mathcal{N}\!\left(0,\sigma_w^2\right),
  \label{eq:ar1_cs3}
\end{equation}

\begin{figure*}
    \centering
    \includegraphics[width=1\linewidth]{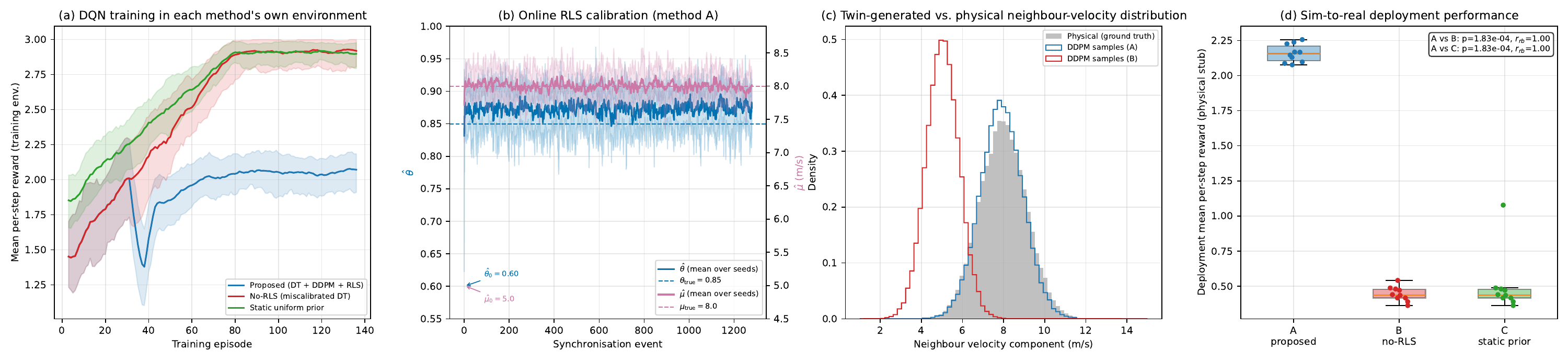}
    \caption{Diffusion-augmented DT case study.
    \textbf{(a)}~DQN training reward in each method's own training
    environment (shaded: min--max over $N_s=10$ seeds).
    \textbf{(b)}~Online RLS calibration of $\hat{\theta}$ (left axis) and
    $\hat{\mu}$ (right axis).
    \textbf{(c)}~DDPM-generated neighbour-velocity distribution of the
    calibrated (A) and miscalibrated (B) twins against the physical ground
    truth.
    \textbf{(d)}~Deployment on the physical stub: per-seed mean per-step
    reward with the Mann--Whitney outcomes of Table~\ref{tab:stats}.}
    \label{fig:cs3_results}
\end{figure*}

with $\theta_{\mathrm{true}}\!=\!0.85$, $\mu_{\mathrm{true}}\!=\!8$\,m/s and $\sigma_w\!=\!0.6$; telemetry is corrupted by measurement noise of standard deviation $\sigma_m\!=\!0.8$. As before, the DT starts deliberately miscalibrated: $\hat{\theta}_0\!=\!0.6$ and $\hat{\mu}_0\!=\!5$\,m/s
(29\% and 38\% error, respectively).
 
\medskip
\noindent\textbf{Online Calibration.}
Every $\Delta_{\mathrm{sync}}\!=\!5$ steps, a telemetry burst is cleaned by the Kalman filters of Eqs.~\eqref{eq:kf_pred}--\eqref{eq:kf_cov} ($Q\!=\!0.3$, $R\!=\!\sigma_m^2\!=\!0.64$) and passed to an online RLS calibrator. A joint regression on $[\,v_t,\;1\,]^{\!\top}$ is nearly collinear for a stationary series and suffers covariance wind-up under forgetting, so Eq.~\eqref{eq:ar1_cs3} is factorised into two persistently excited scalar stages. Stage~1 tracks the stationary mean,

\begin{equation}
\begin{aligned}
k^{\mu} &=
\frac{P^{\mu}}{\lambda_f + P^{\mu}},\\
\hat{\mu} &\leftarrow
\hat{\mu}
+ k^{\mu}\!\left(v_{t+1}-\hat{\mu}\right),\\
P^{\mu} &\leftarrow
\frac{(1-k^{\mu})P^{\mu}}{\lambda_f}.
\end{aligned}
\label{eq:rls_cs3_gain}
\end{equation}
and Stage~2 estimates the persistence on the mean-centred pair
$\phi_t = v_t - \hat{\mu}$, $y_t = v_{t+1} - \hat{\mu}$:
\begin{align}
  k^{\theta} &= \frac{P^{\theta}\phi_t}
                 {\lambda_f + \phi_t^2 P^{\theta}},\qquad
  \hat{\theta} \leftarrow \hat{\theta} +
                 k^{\theta}\!\left(y_t - \phi_t\,\hat{\theta}\right),
  \label{eq:rls_cs3_update}\\
  P^{\theta} &\leftarrow \frac{1}{\lambda_f}
                 \left(1 - k^{\theta}\phi_t\right)P^{\theta},
  \label{eq:rls_cs3_cov}
\end{align}
with $\lambda_f\!=\!0.995$, updates pooled over all 12 neighbour-velocity components, and both estimates clipped to plausible ranges. A z-score residual monitor identical to Eq.~\eqref{eq:anomaly} runs in parallel.
 
\medskip
\noindent\textbf{Diffusion Model inside the Twin.}
The DT hosts a conditional denoising diffusion probabilistic model (DDPM) that generates the neighbour-velocity context
$\mathbf{x}_0 \in \mathbb{R}^{12}$ conditioned on the ego velocity: the forward process corrupts a twin-simulated sample over $T\!=\!50$ steps, $q(\mathbf{x}_t \mid \mathbf{x}_0) =
\mathcal{N}(\sqrt{\bar{\alpha}_t}\,\mathbf{x}_0, (1-\bar{\alpha}_t)\mathbf{I})$, an MLP denoiser $\boldsymbol{\epsilon}_\vartheta(\mathbf{x}_t, t, \mathbf{c})$ learns to invert it, and ancestral sampling yields synthetic neighbour states. Its training data are rollouts of the twin's \emph{own calibrated
dynamics}, refreshed three times as calibration progresses, so the samples converge in distribution to the physical swarm as $\hat{\theta} \to \theta_{\mathrm{true}}$. This replaces the static uniform-noise assumption of the underlying implementation, whose regression-style generator collapses to the conditional mean.
 
\medskip
\noindent\textbf{Reward and Policy.}
At each step, a DQN, whose state concatenates the ego velocity with the DDPM-generated neighbour velocities, selects $v^{\mathrm{own}}_t$. The reward couples separation with a mission-speed incentive,
\begin{equation}
  r_t \;=\; \sum_{j=1}^{J}
  \mathbf{1}\!\left[\;\bigl\lVert \mathbf{v}^{(j)}_t -
  v^{\mathrm{own}}_t\mathbf{1} \bigr\rVert_2 > d_{\mathrm{safe}}\right]
  \;-\; \lambda_v \bigl| v^{\mathrm{own}}_t - v_{\mathrm{mission}} \bigr|,
  \label{eq:reward_cs3}
\end{equation}
with $d_{\mathrm{safe}}\!=\!3$, $v_{\mathrm{mission}}\!=\!8$\,m/s and $\lambda_v\!=\!0.3$. The penalty rules out fleeing to an extreme velocity; the remaining trade-off between formation-keeping and separation is resolved correctly only if the twin's belief about the neighbours is correct, which is exactly what the calibration loop provides.
 
\medskip
\noindent\textbf{Pipeline and Baselines.}
The four-phase pipeline of Section~\ref{sec:casestudy} is retained: the DQN trains only inside the DT (140 episodes of 40 steps) while telemetry calibrates the twin in parallel (1280 synchronisation events per run); the greedy policy is then deployed unchanged on the physical stub, where synchronisation continues. Two ablations isolate each component: \emph{no-RLS} (B), with $\hat{\theta},\hat{\mu}$ frozen at their erroneous initial values, and \emph{static prior} (C), with neighbours drawn uniformly from $[1,15]^{12}$ as in the original implementation.
 
\medskip
\noindent\textbf{Statistical Analysis.}
Each method is trained with $N_s\!=\!10$ independent seeds and evaluated greedily on the physical stub (20 episodes of 40 steps); the per-seed metric is the mean per-step deployment reward. Since these scores are few and not guaranteed to be normally distributed, methods are compared with the non-parametric Mann--Whitney $U$ test (proposed vs.\ each baseline). Because two simultaneous comparisons are made, a Bonferroni correction reduces the significance threshold to $\alpha' = 0.05/2 = 0.025$. Effect size is quantified via the rank-biserial correlation
\begin{equation}
    r_{rb} \;=\; 1 - \dfrac{2U}{n_1 n_2},
    \label{eq:rrb}
\end{equation}
where $U$ is the Mann--Whitney statistic and $n_1 = n_2 = N_s$; conventional thresholds are $|r_{rb}| < 0.30$ (small), $0.30$--$0.50$ (medium), $> 0.50$ (large). Table~\ref{tab:stats} reports all outcomes.
 
\begin{table}[t]
\caption{Statistical comparison of deployment performance (mean per-step reward on the physical stub, $N_s=10$ seeds per method). Mann--Whitney $U$ tests with Bonferroni-corrected threshold $\alpha' = 0.025$; effect size is the rank-biserial correlation of Eq.~\eqref{eq:rrb}.}
\label{tab:stats}
\centering
\scriptsize
\setlength{\tabcolsep}{3pt}

\begin{tabular}{|p{2.45cm}|c|c|c|c|c|c|}
\hline
\textbf{Comparison} &
\shortstack{\textbf{Med. A}\\\textbf{[IQR]}} &
\shortstack{\textbf{Med. Base.}\\\textbf{[IQR]}} &
$\mathbf{U}$ &
$\mathbf{p}$ &
$\mathbf{r_{rb}}$ &
\textbf{Sig.} \\
\hline
A vs.\ B (no RLS)
& 2.16 [0.11]
& 0.44 [0.06]
& 0
& $1.83\times10^{-4}$
& 1.00
& Yes \\
\hline
A vs.\ C (static prior)
& 2.16 [0.11]
& 0.44 [0.06]
& 0
& $1.83\times10^{-4}$
& 1.00
& Yes \\
\hline
\end{tabular}

\vspace{2pt}
\begin{minipage}{\columnwidth}
\scriptsize
\textit{Notation:} Med. = median; Base. = baseline; IQR = interquartile range; $U$ = Mann--Whitney $U$ statistic; $p$ = $p$-value; $r_{rb}$ = rank-biserial correlation; Sig. = statistical significance.
\end{minipage}

\end{table}

\medskip
\noindent\textbf{Results.}
Fig.~\ref{fig:cs3_results} summarises the study. RLS drives $\hat{\theta}$ from $0.6$ to $0.875$ and $\hat{\mu}$ from $5$ to $8.01$\,m/s, within 2.9\% and 0.2\% of the true values, using only noisy telemetry (panel~(b)), and the DDPM trained on the calibrated twin reproduces the physical velocity distribution, while the no-RLS generator stays centred on the wrong mean (panel~(c)). Panel~(a) exposes the classic sim-to-real trap: both baselines earn \emph{higher} training reward inside their own (wrong) twins, which overestimate how easy separation is. Deployment reverses the picture (panel~(d)): the proposed method reaches a median per-step reward of $2.16$ versus $0.44$ for both baselines, significant at the corrected threshold ($p=1.83\times 10^{-4}$ for both comparisons) with maximal effect sizes ($r_{rb}=1.00$). The residual monitor flagged only 0.2\% of post-burn-in samples, consistent with its three-sigma design (Eq.~\eqref{eq:anomaly}). As in Section~\ref{sec:casestudy}, the physical layer is emulated by a software stub; hardware-in-the-loop validation is left as future work.

\subsection{Discussion}
By integrating DTs with DRL, a virtual replica of the UAV and its environment enables realistic simulations that provide a high-fidelity virtual model for real-time monitoring and optimization. This integration significantly improves the applicability of DRL by removing the key limitations of traditional RL approaches. DT-based DRL frameworks overcome the limitations of the RL exploration phase and
enable more efficient and effective learning processes. In addition, these frameworks address the practical challenges associated with DRL applications, such as reducing computational overhead and improving deployment efficiency. As demonstrated in Section~\ref{sec:casestudy}, online RLS calibration can close a substantial initial model mismatch, 28\% in the path-loss exponent, passively, as a by-product of routine synchronisation events, and the resulting policy transfers directly to the physical layer without any architectural change.
\par
One of the key merits of DT-based DRL is its ability to reduce training time while balancing training speed and environmental fidelity.  DTs create a detailed and accurate virtual environment, allowing DRL models to reach a high level of performance without relying on extensive real-world trials. The potential of DTs can be further enhanced by advances in ML, particularly through the integration of  GAI. Large Language Models (LLMs) bring a new dimension to DTs \cite{10591881} by enabling intuitive interactions in natural language. This allows operators to interact seamlessly with DT systems, reducing the need for specialized expertise. LLMs also improve predictive analytics by leveraging historical and real-time data to make accurate predictions and support decision-making in UAV networks. Techniques such as Retrieval Augmented Generation enable LLMs to incorporate external knowledge, enhancing the contextual and analytical capabilities of DTs\cite{xie2026trustworthy},\cite{xiong2025dr}. In addition, the dynamic fine-tuning and continuous learning capabilities of LLMs enable DTs to effectively adapt to evolving UAV scenarios and ensure robust and context-aware decision-making.
\par

\subsection{Key Findings and Insights}

Based on the analysis in the preceding subsections, it is evident that DT can improve UAVs' navigation and path planning and promote UAV swarms' performance. In this section we outline key issues where further developments are expected:

\subsubsection{Support of 5/6G Technologies}
5G networks are reshaping how DTs are developed and utilized in UAV networks. Leveraging ultra-low latency and high data rate, real-time UAV swarm data can be integrated into their models, resulting in more precise and timely DT. This advancement, powered by 5G, enhances optimization outcomes by enabling higher data rates, real-time monitoring, efficient collaboration, and improved security measures. Consequently, 5G is driving significant innovation in DT technology, facilitating smarter and more responsive UAV systems \cite{dzone_2023}. In this context, Kunst \textit{et al.} \cite{9283775} evaluated the performance of 4G, 5G, and 6G networks in terms of three QoS metrics essential for real-time applications: latency, delay, and jitter. Regarding latency, the results demonstrate that 6G reduces latency by up to four times compared to 5G and up to six times compared to 4G. For delay, the 6G results are approximately three times better than those achieved with 5G. Regarding jitter, 6G is expected to outperform the other technologies, maintaining the average jitter below the specified limit in scenarios with up to 1,200 UAVs. In contrast, 4G and 5G exhibit similar behaviors, supporting QoS-enabled communication for up to 200 and 300 UAVs, respectively. 

\subsubsection{The Impact of Varying Levels of Autonomy}

The varying levels of UAV autonomy are listed as follows:
\begin{itemize}
    \item Level 1 - Remotely Controlled System:  
   UAVs are fully controlled by operators, with no autonomous behavior. The system reactions depend entirely on human input.
   \item Level 2 - Automated System:
   UAVs perform predefined tasks based on built-in functionality or programming, with limited autonomy and no real-time adaptation.

   \item  Level 3 - Autonomous Non-Learning System:  
   UAVs can perform goal-directed tasks based on fixed rules, with limited real-time decision-making and the ability to re-plan missions in response to changing conditions.

   \item Level 4 - Autonomous Learning System:
   UAVs can modify their behavior based on real-time learning, adapting to dynamic environments and continuously improving their decision-making abilities while following safety protocols~\cite{protti2007uav}.
\end{itemize}
The level of autonomy significantly influences the requirements of DT-based solutions. Low Autonomy (Levels 1..2): DT models can focus on static simulations, fixed trajectories, and basic fault management, supporting limited automation with human-in-the-loop operations. High Autonomy (Levels 3..4): DTs must provide real-time synchronization, advanced simulation environments, and adaptive and intelligent decision-making capabilities based on AI. 

\begin{table}[htbp]
\centering
\caption{Taxonomy of Synchronization Strategies for UAV Digital Twins}
\label{tab:synchronization_taxonomy}

\begin{tabular}{|p{0.5cm}|p{2cm}|p{5.2cm}|}
\hline
\textbf{Ref} & \textbf{Category} & \textbf{Description} \\
\hline

\cite{10190734} & Real-time bidirectional synchronization &
Physical UAV and DT continuously exchange data; observations and actions from agents are sent to the DT layer to update virtual twins in real time. \\
\hline

\cite{10190734} & State-based synchronization (predictive) &
DT dynamically estimates key hyperparameters (e.g., task information distribution) in real time and uses them to guide decision-making. \\
\hline

\cite{9614346} & Centralized DT training with physical-digital connection &
DT is incorporated into a central system to train the DRL model; after training, the model is transferred to real UAVs by connecting physical and digital units. \\
\hline

\cite{zhou2024hierarchical} & Hierarchical / edge-aggregated synchronization &
Edge UAVs aggregate lightweight DT models derived from local physical information with low computational cost; real-time exchange of parameters between regions occurs. \\
\hline

\cite{10076786} & Twin data layer bridging &
A dedicated twin data layer (using MySQL database) bridges physical and digital spaces, facilitating data collection, monitoring, control, and fault reproduction in real time. \\
\hline

\cite{cao2025uav} & Twin training + continuous evolution &
A DT-driven training framework where synchronization is part of a closed loop: twin training, decentralized execution, and continuous evolution. \\
\hline

This work & Kalman-filtered bidirectional sync with RLS parameter update & Every $\Delta_{\mathrm{sync}} = 5$ decision steps, noisy UAV telemetry (GPS position, per-sensor path-loss) is received and cleaned via a bank of scalar Kalman filters. The filtered values feed an online RLS estimator that updates the DT's path-loss exponent $\hat{\alpha}$ in real time. A z-score residual test running in parallel flags anomalies consistent with GPS spoofing or FDI attacks. \\
\hline
\end{tabular}
\end{table}

\subsubsection{Computational Overhead}

The computational overhead and associated costs are key factors in implementing DTs for UAV applications. Developing ultra-high-fidelity models of systems and simulating their complex processes in real-time requires significant computing power, advanced software, skilled expertise, and extensive data processing capabilities. These processes are time and labor-intensive, resulting in significant development and operational costs. Consequently, the high cost of DT implementation for UAV operations can be a barrier to their adoption, especially for smaller UAV networks~\cite{asi4020036}. DMs can generate synthetic data samples that closely match the real-world distribution of data without the need to perform computationally expensive operations. DMs can sample from the learned distribution to generate realistic data for the DT. This reduces the computational resources needed for frequent data collection or simulations\cite{emami2025diffusion}.

\subsubsection{Data Synchronization}

Data synchronization in DTs ensures accurate, real-time alignment between the physical and virtual systems. This process involves collecting large amounts of sensor and operational data from UAVs and transferring it to the digital model while ensuring minimal latency and high data fidelity. Dealing with heterogeneous data from different sources and maintaining consistent updates despite possible transmission interruptions are two salient challenges that impact effective synchronization~\cite{electronics13071263}, thereby controlling them facilitates real-time monitoring and error detection and improves decision-making and predictive capabilities essential for UAV dynamic environments. DMs can be integrated into a DT to improve data transmission and synchronization. The DM is trained offline to handle data compression, noise reduction, and signal reconstruction tasks, enabling efficient and reliable communication between the DT and UAV. By acting as a generative transmitter and receiver, the DM compresses critical information for transmission and restores it to its original state upon arrival, even over noisy channels. This approach ensures real-time synchronization between the UAV and its DT enhancing operational efficiency and supporting seamless coordination in dynamic UAV environments\cite{emami2025diffusion}. Table \ref{tab:synchronization_taxonomy} provides a taxonomy of synchronization strategies.

\subsubsection{Cybersecurity}
DT technology can significantly improve cybersecurity~\cite{10263803} by enabling real-time monitoring and anomaly detection in systems \cite{Homaei2024}. By creating a high-fidelity virtual replica of the UAV and its environment, DT allows for continuous tracking of the UAV's operational state, including flight parameters, sensor data, and communications. This virtual model can be used to detect discrepancies or anomalies in real-time, such as those caused by cyberattacks like GPS spoofing. In this approach, DT models can be integrated with data-driven methods and ML techniques to identify cyber intrusions. For example, by applying novelty detection on flight data, DT can flag unusual patterns or behaviors that deviate from expected norms, signaling potential cyber threats. Multiple ML models, including classical and DL methods, can be trained using the DT architecture to optimize detection accuracy and responsiveness. These models help identify the most effective techniques for recognizing modern cyber-intrusions, such as spoofing, jamming, or unauthorized access before they can impact the UAV’s operation~\cite{9594321}. By continuously updating the digital replica with real-world data, DT improves the UAV's ability to recognize and respond to emerging cybersecurity threats in real-time, ensuring robust protection against attacks. This proactive approach enables better prevention, detection, and mitigation of cyber risks, enhancing the overall security and reliability of UAV systems.

\subsubsection{DT for AI Development, Simulation, and Real-World Validation
} 
Simulation tools such as Gazebo, Unity, and AirSim provide convenient platforms for developing and testing DT models. However, real-world validation remains indispensable to ensure the reliability and accuracy of these systems. Simulations may fail to account for unpredictable environmental factors like weather and terrain, which can significantly impact UAV behavior. To address this, testbed deployments and field testing of DT frameworks on physical UAVs are often employed, helping confirm the performance and operational feasibility of these technologies in real-world conditions. These validations are crucial for assessing the effectiveness of DT systems in enhancing UAV situational awareness and control, ensuring they deliver accurate, real-time feedback during missions \cite{10556896}. Building on this, Gurses \textit{et al.} \cite{gurses2404digital} presented a case study on AI-assisted signal source localization, showcasing the full lifecycle of AI algorithm development, testing, and validation within the DT environment of the AERPAW platform. This seamless integration extended to real-world testing, demonstrating DTs as an effective platform for practical AI development in autonomous vehicle networks. Experimental results compared three workflows: development solely in DT, DT-based development with uncalibrated real-world testing, and DT development calibrated with real-world data. The calibrated approach delivered superior performance, underscoring the value of real-world data in refining AI algorithms within DT frameworks. DTs significantly enhance the development and deployment of real-world UAV applications by creating a seamless, bidirectional bridge between virtual simulation and physical operation. They function as a hybrid testbed where AI algorithms for computer vision, swarming, and control can be rigorously trained and validated in a risk-free, customizable virtual environment before being deployed to real UAVs. This process ensures robustness, as algorithms are tested against complex, simulated scenarios that are difficult or unsafe to replicate physically, such as dynamic surveillance missions. Moreover, the DT enables critical evaluation of different system architectures, assessing whether computation is best handled on-board the drone, at the network edge, or in the cloud—and analyzes the impact of communication technologies like 5G on latency and performance. By providing a realistic, repeatable, and safe prototyping platform, DTs drastically reduce development costs and logistical challenges, leading to more reliable and effective UAV deployments in critical real-world applications like security, defense, and search-and-rescue\cite{carraminana2024digital}.

\subsubsection{Real World Deployment of DT-XR Integration}

The integration of VR and DT technologies for UAV operations combines immersive interaction and real-time virtual replication to enhance control, monitoring, and situational awareness. Utilizing VR Head-Mounted Displays (HMDs) and controllers, operators interact with a DT environment that mirrors the UAV's physical system. Real-time data transmission, enabled by 5G networks, synchronizes UAV states across virtual and physical systems. This integration offers applications in emergency response, surveillance, and e-commerce. A promising development in the effectiveness of DT-XR integration is offered in \cite{10004244}. This system consists of a DT with two parts: a virtual environment and the actual physical UAVs. The VR setup includes an HMD that provides the operator with a DT of the UAV and its surroundings. The operator uses VR controllers to pilot the UAVs, with control data being transmitted to the real UAVs in real-time. Experimental results show this VR-based system offers a stable, intuitive, and immersive control method. It allows for ultra-remote control and significantly improves the ability to manage multiple UAVs simultaneously.

\subsubsection{Data Management}
DRL models require extensive training data to develop and deliver accurate predictions. However, collecting sufficient and representative data can be particularly challenging for systems with limited historical data. Operators may need to use data augmentation or synthetic data generation techniques to address data scarcity. The combination of DM and DT allows for the creation of diverse synthetic data that accurately reflects real-world dynamics, which can be used to train RL models more effectively\cite{emami2025diffusion}. In addition, it is important to ensure the accuracy, consistency, and reliability of the data used by the DRL models. This requires operators to validate and maintain the quality of data collected from sensors and other sources. Furthermore, the seamless flow of data between DTs and DRL models requires the implementation of robust data integration mechanisms to support efficient and effective training and decision-making~\cite{9359733}. The data island problem in traditional AI arises from the isolation of valuable datasets across devices due to privacy concerns and regulatory restrictions, limiting the effectiveness of centralized AI models. Federated Learning (FL) addresses this challenge by enabling decentralized, privacy-preserving model training across distributed devices, and its integration with DTs offers a promising solution to overcome data silos. However, FL faces critical issues such as single-point failures, susceptibility to poison attacks, and the lack of effective incentive mechanisms~\cite{homaei2026feature}. To tackle these challenges, blockchain technology has been proposed as a complementary solution, providing a decentralized, secure, and immutable ledger to enhance FL’s robustness, trustworthiness, and fairness, thereby facilitating the successful deployment of DTs\cite{LIU2024248}.

\subsubsection{Regulatory Frameworks}
The integration of DT technology into UAV operations offers a transformative pathway to achieving regulatory compliance, safety, and certification efficiency. DTs serve as powerful enablers for data-driven oversight. By creating a dynamic virtual replica of a UAV and its operational environment, DTs allow continuous monitoring, simulation, and risk assessment across all phases of a UAV’s lifecycle. This capability directly supports the specific operations risk assessment process by enabling virtual validation of design robustness, operational safety, and airworthiness without relying solely on costly or time-consuming flight tests. However, realizing their full regulatory potential demands standardized data models, secure interoperability, and clear legal recognition of DT-based certification evidence. DTs are poised to become a cornerstone technology linking engineering innovation with regulatory accountability, ensuring that future UAV operations are not only efficient but demonstrably safe and compliant\cite{fakhraian2023towards}.

\subsubsection{Environment and Multiple Tasks}
Designing DTs that remain robust across different environments adds another layer of complexity. UAVs operate in rapidly changing environments, requiring DTs to adapt and respond to real-time changes. Also, managing interactions with environmental objects and other UAVs is essential to maintain smooth operations. UAVs may need to perform multiple tasks simultaneously or coordinate with other UAVs or environmental objects. DTs capable of managing such complexities and interactions effectively require advanced modeling, simulation capabilities, and integration with ML~\cite{10089851}. 

\subsubsection{Millimeter-wave Communications}
5G communication links ensure reliable and low latency communications with improved capacity, thereby appearing as a practical option for communication between UAVs and DTs. Millimeter-wave (mmWave) offers a broader range of frequency resources compared to the microwave band, enabling significantly higher achievable rates to support UAV services such as video surveillance, hotspot coverage, and emergency communications~\cite{9205984}. Also, mmWave communications frequencies have great potential to fulfill the requirements for future high-speed communication links between UAVs and DTs. The UAVs can conduct flexible mmWave beamforming to mitigate the high path loss caused by the long distance.

\subsubsection{Ethical Issues}
In a scenario where each UAV has a DT and one of them fails, there are some important considerations for the transfer of the previous UAV's experience to the new UAV's DT. Ideally, a new DT could take over the data or experience of the DT of the failed UAVs via cloud-based or distributed systems so that it can start with the same operational data and learning. However, issues such as data ownership, privacy, and security arise, especially when considering the ethical implications of transferring personal or proprietary experience from one UAV to another\cite{9076112}.

\section{Extended Reality Revolutionizing UAVs: Navigating Paths and Enhancing Awareness} \label{sec3}

This section examines the transformative role of XR technologies in UAVs. We examine the role of XR in path planning and navigation and highlight its role in training and control. Finally, we examine the contributions of XR to situational awareness, collision avoidance, safety, and surveillance. Fig. \ref{fig:uav_layer} highlights XR contributions for UAVs. XR for UAVs enhances situational awareness by providing immersive visualizations, improves operator training through realistic simulations, and increases safety with augmented real-time data. It boosts mission success rates with better planning, aids in navigation and collision avoidance using detailed overlays, and supports efficient decision-making with intuitive interfaces and actionable insights.

\begin{figure*} [h]
    \centering 
    \captionsetup{justification=raggedright}
    \includegraphics[width=18cm,height=10cm]{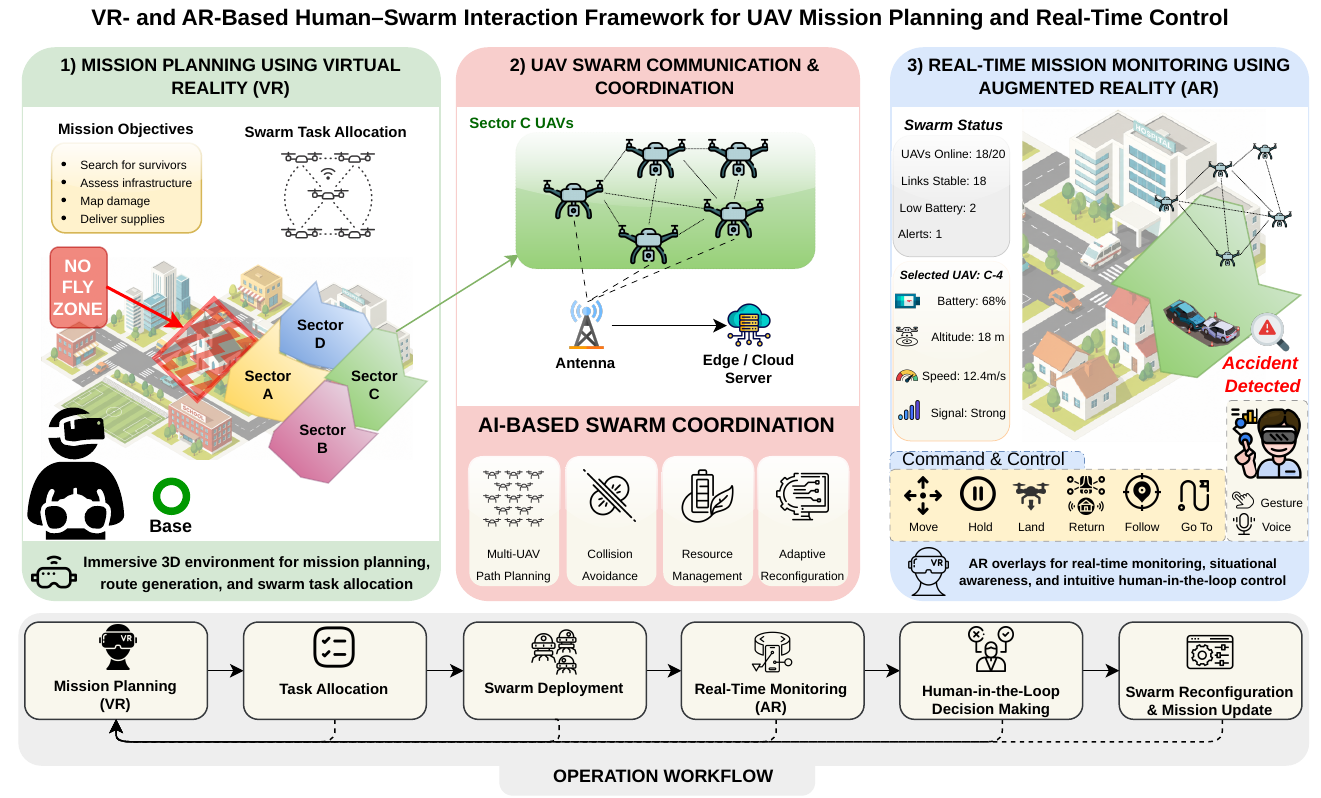}
    \caption{Conceptual framework of a VR- and AR-enabled UAV swarm system for disaster response. Virtual Reality provides an immersive environment for mission planning, path generation, and swarm task allocation before deployment. During flight, Augmented Reality overlays real-time telemetry, UAV positions, sensor information, and detected targets onto the operator's view, enabling intuitive monitoring and human-in-the-loop control for dynamic mission adaptation.}
    \label{fig:digital}
\end{figure*}

\par
\begin{table}[h]
\caption{Key Techniques for AR and VR Technologies in UAV Navigation and Path Planning}
\label{tab30}
\centering
\scriptsize
\begin{tabular}{|p{0.5cm}|p{1.75cm}|p{1.75cm}|p{3.2cm}|}
\hline
\textbf{Ref} & \textbf{Technique} & \textbf{Application} & \textbf{Strengths and Limitations} \\
\hline
\cite{MOURTZIS2022983} & AR-based Interface & Path planning in indoor facilities & Minimizes risk of errors and collisions; aids UAV manipulation in complex environments. \\
\hline
\cite{10.1145/3357251.3362742} & VR-based Platform & 3D flight path planning & Improves safety over manual control; more efficient than 2D interfaces. \\
\hline
\cite{8942250} & VR with Web Technology & Navigation and path planning & Enables practical knowledge; real-world data for accurate path planning. \\
\hline
\cite{8243463} & UAV DL Virtual Reality & Training data collection and route planning & Generates large amounts of training data; improves AI training efficiency. \\
\hline
\end{tabular}
\end{table}

\subsection{Literature Review}
\subsubsection{Path Planning and Navigation}

In addressing challenges in UAV path planning and object location, Mourtzis \textit{et al.}~\cite{MOURTZIS2022983} advocated for an AR-based user interface tailored for indoor manufacturing facilities. This interface minimizes errors and collision risks by enhancing spatial awareness and facilitating intuitive UAV control amidst obstacles. In another domain, Paterson \textit{et al.}~\cite{10.1145/3357251.3362742} presented an open-source platform for 3D flight path planning in VR and compare its performance with existing UAV pilot interfaces. VR technology enhances safety compared to manual control and efficiency compared to a 2D touchscreen. Furthermore, Nguyen \textit{et al.}\cite{8942250} developed a VR technology mixed with web technology to address UAVs navigation, path planning, and crash probability. It enables operators to gain practical knowledge of UAVs before they are deployed. The VR environment was reconstructed based on real-world airborne data and path planning and navigation were performed.
\par
Moreover, Wang \textit{et al.}\cite{8243463} used the UAV virtual platform to collect a large amount of training data and images, which can effectively allow UAV route planning and self-control in the UAV flight process cost-effectively. The developed UAV DL virtual reality system can generate a large amount of training data and improve the training efficiency of AI systems. Table~\ref{tab30} summarizes the discussed content.
\par

Existing DT platforms for UAV environments focus primarily on modeling, prediction, and simulation, but lack rigorous evaluation of real‑time data communication, a critical requirement for bidirectional control between virtual and physical spaces, especially for DT maturity levels 3 and 4. Bae \textit{et al.}\cite{bae2025digital} introduce RC‑DT‑Platform, a DT platform specifically designed and evaluated for real‑time data communication in UAV operations. Experimental results demonstrate that RC‑DT‑Platform can transmit approximately 454 data/sec for 100‑byte messages, 119 data/sec for 100‑KB messages, and 0.7 data/sec for 16‑MB messages. With 32 registered objects, the platform achieves a read throughput of about 3500 data/sec, independent of data size. Compared to the baseline Ditto framework, RC‑DT‑Platform degrades by less than 6.25× as registered objects increase, whereas Ditto degrades by up to 10×. Notably, the proposed platform achieves up to 46× higher read performance and up to 12× higher write performance than Ditto. This work fills a critical gap by validating that real‑time data communication is achievable and by providing a performance baseline for future mission‑critical UAV DT deployments.

\begin{table}[h!]
\caption{Key Techniques for XR Technologies in UAV Control and Training}
\label{tab40}
\centering
\scriptsize
\begin{tabular}{|p{0.5cm}|p{1.75cm}|p{1.75cm}|p{3.2cm}|}
\hline
\textbf{Ref} & \textbf{Technique} & \textbf{Application} & \textbf{Strengths and Limitations} \\
\hline
\cite{4741347} & VR-based Environment & UAV control and pilot training & Merges UAV camera images with geographic data; overcomes control method disadvantages. \\
\hline
\cite{10004244} & VR-based Approach & Multi-person UAV swarm control & Uses VR controllers and head-mounted displays; operates with low-latency 5G; solves network resource allocation issues. \\
\hline
\cite{s21041456} & AR Application & UAV pilot training and evaluation & Facilitates learning process; enriches images with virtual content. \\
\hline
\cite{liu2018usability} & VR Interfaces & Control and mission planning & Evaluates usability; experiments with new interaction methods; focuses on onboarding experience. \\
\hline
\end{tabular}
\end{table}
\subsubsection{UAV control and training}
As described in what follows, XR has been used for UAV control and training.
Zhi-Hua \textit{et al.}~\cite{4741347} addressed the deficiencies of the UAV control method and developed a virtual flight environment based on VR technology to resolve the problems of training UAV pilots. VR technology assists in merging dynamic images from the UAV camera with the geographic data information generated by the ground computer. Similarly, Ribeiro \textit{et al.}~\cite{s21041456} developed a new AR application for UAV pilot training and evaluation to facilitate the learning process of UAV piloting. In this application, the user can make informed decisions about controlling a UAV based on images captured by the camera and enriched with virtual content displayed on a screen. In addition, Chen \textit{et al.}~\cite{10004244} developed a VR-based approach to multi-person UAV swarm control that can operate simultaneously with low-latency 5G infrastructure. The authors used VR controllers and head-mounted displays to control UAVs, solving the problem of multiple UAVs competing for network resource allocation. Liu \textit{et al.}~\cite{liu2018usability} assessed the usability of VR interfaces for controlling UAVs by collaborating with UC Berkeley’s Immersive Semi-Autonomous Aerial Command System (ISAACS) project. Their experiments explored innovative human-UAV interactions in a VR environment, focusing on the UAV mission planning phase and developing an onboard experience for new users. 

Vona \textit{et al.}\cite{vona2025unmann} shows that the choice of virtual point of view (POV) in a DT–based VR UAV control system has a direct impact on both pilot performance and user experience. While First Person View (FPV) improves immersion and leads to smoother control inputs, it also increases cognitive workload and reduces flight precision. In contrast, Third Person View (TPV) and Chase View generally reduce mental demand and improve overall control accuracy, with TPV emerging as the most preferred perspective among participants. These results suggest that no single POV is universally optimal; instead, effective UAV teleoperation systems should adapt the viewpoint to balance situational awareness, usability, and task complexity.
Table \ref{tab40} summarizes the discussed content.

\subsubsection{Situational Awareness}
Ruano \textit{et al.}~\cite{s17020297} proposed the AR system with two functionalities to improve situational awareness of UAVs with medium altitude and long endurance in reconnaissance missions. The route orientation function is responsible for predicting the next waypoint and the further path of the UAV. The target identification function is responsible for quickly locating the target. In a related vein, Dorzhieva \textit{et al.}~\cite{9995060} proposed DRL-based collision avoidance and UAV swarm control for an environment with hovering objects, along with a new AR technology with a wearable tactile device for human-UAV interaction. The visualized trajectory and haptic feedback allow the user to anticipate the UAV's path. DRL approach makes interaction safe. Similarly, Ji \textit{et al.}~\cite{8068132}, developed a data-driven UAV ground station, AR display, and control system, and evaluated their performance in a quadrotor flight experiment. The developed AR technology uses the UAV telemetry data and video images and merges them to allow ground station operators to observe real-time information about the aircraft attitude, position, and operational status. 

\begin{table}[htbp]
\centering
\caption{Taxonomy of Human-Robot Interaction (HRI) Modalities for UAVs in XR Environments}
\label{tab:hri_modality_taxonomy}

\begin{tabular}{|p{0.5cm}|p{1.8cm}|p{5.4cm}|}
\hline
\textbf{Ref} & \textbf{Category} & \textbf{Description} \\
\hline
\cite{10004244} & VR controllers + head-mounted display (HMD) & Handheld VR controllers and HMD used to pilot UAVs; low-latency 5G enables multi-person swarm control. \\
\hline
\cite{9341037}  & AR head-worn device (HoloLens) & Mixed reality headset that reconstructs and displays a 3D map of UAV surroundings; user can edit and render the map interactively. \\
\hline
\cite{s21041456},\cite{MOURTZIS2022983}   & AR screen / tablet & Augmented reality overlays on a 2D screen; enriches UAV camera images with virtual content for pilot training and indoor path planning. \\
\hline
\cite{9995060} & Wearable tactile / haptic device & A tactile device worn by the user provides haptic feedback (e.g., visualized trajectory) to anticipate UAV path; combined with DRL for collision avoidance. \\
\hline
\end{tabular}

\end{table}

Traditional radio-based information exchange between humans during emergency operations suffers from a lack of visualization, often leading to miscommunication. Furthermore, Agrawal \textit{et al.}~\cite{ankit} used the location-based AR paradigm to geotag, share, and visualize information. This paradigm aims to support bidirectional communication between humans and UAVs and to visualize information about the scene relevant to the rescue team's role. Additionally, Liu \textit{et al.}~\cite{9341037} developed a new AR interface based on Microsoft HoloLens 1. 
HoloLens is a head-worn mixed reality device that interacts with the autonomous UAV. The UAV can reconstruct a 3D map showing its surroundings. This map is then reflected on the AR interface and can be edited and rendered by the user. Moreover, Llasag \textit{et al.}~\cite{8760811} addressed human detection in natural disaster areas using UAVs. An architecture for search and rescue missions uses a mixed-reality interface to improve location determination and video streaming. Table~\ref{tab50} summarizes the discussed content. 

In the context of security, Yang \textit{et al.}~\cite{10195191} addressed information leakage in UAV-based AR systems with wireless communication. They designed a UAV-based covert communication system in which a UAV transmits a covert message to receivers on the ground while a cooperative UAV acts as a jammer to interfere with malicious eavesdroppers. This system ensures the protection of private information for cooperative AR users and allows them to transmit covert messages without being monitored. Table \ref{tab:XR_comparison} provides a comparative overview of recent research on the use of XR technologies, both AR and VR, for UAV navigation, control, and training. The studies focus on diverse objectives, including indoor path planning, collision avoidance, multi-UAV swarm coordination, and pilot training. AR applications are often employed for enhancing spatial awareness and human–UAV interaction, sometimes incorporating haptic feedback, while VR is used for immersive flight simulations, multi-user control, and reconstruction of real-world environments to improve realism. Across the studies, XR technologies are shown to improve safety, situational awareness, skill transfer, and interaction feasibility, particularly in multi-user or networked scenarios. Overall, the table illustrates that XR offers significant potential to enhance UAV operations and training, though practical deployment challenges remain a key consideration. Table \ref{tab:hri_modality_taxonomy} provides a taxonomy of human-robot interaction modalities.

\begin{table}[t]
\caption{Key Techniques for AR and DRL-based Technologies in UAV Control and Situational Awareness}
\label{tab50}
\centering
\scriptsize
\begin{tabular}{|p{0.5cm}|p{1.75cm}|p{1.75cm}|p{3cm}|}
\hline
\textbf{Ref} & \textbf{Technique} & \textbf{Application} & \textbf{Strengths and Limitations} \\
\hline
\cite{s17020297} & AR System & Situational Awareness & Improves situational awareness; route orientation predicts waypoints; target identification locates targets. \\
\hline
\textbf{\cite{8068132}} & Data-driven Ground Station and AR Display & UAV Control & Real-time UAV telemetry and video information; improves situational awareness. \\
\hline
\textbf{\cite{ankit}} & Location-based AR & Emergency Operations & Geotags and visualizes information; supports bidirectional communication. \\
\hline
\textbf{\cite{9341037}} & AR Interface with HoloLens 1 & UAV Control & 3D map reconstruction and display; user interaction and editing. \\
\hline
\textbf{\cite{8760811}} & Mixed-Reality Interface & Search and Rescue & Improves location determination and video streaming. \\
\hline
\cite{10195191} & UAV-Aided Covert Communication & Information Leakage Prevention & Covert communication to protect private information; uses UAVs for transmission and jamming. \\
\hline
\end{tabular}
\end{table}
\begin{table*}[htbp]
\centering
\caption{Comparative analysis of XR frameworks for UAV navigation, control, and training}
\label{tab:XR_comparison}
\footnotesize  
\begin{tabularx}{\textwidth}{@{}l>{\raggedright\arraybackslash}X>{\raggedright\arraybackslash}Xl>{\raggedright\arraybackslash}X>{\raggedright\arraybackslash}X>{\raggedright\arraybackslash}X@{}}
\toprule
\textbf{Ref.} & \textbf{Objective/Problem} & \textbf{XR Technology/Simulator} & \textbf{Validation} & \textbf{Metrics} & \textbf{Strengths} & \textbf{Limitations} \\
\midrule
{\cite{MOURTZIS2022983} } & AR interface for indoor UAV path planning & AR application (industrial demo) & Case study & Error reduction (qualitative) & Improves spatial awareness; reduces collisions & Latency/bandwidth not measured; scalability unclear \\
\midrule
\cite{10.1145/3357251.3362742}  & 3D VR interface for UAV path planning and control & VR flight simulation platform & User study & Task completion time, usability & Safer/faster path planning; enhanced awareness & Motion-sickness/latency not quantified \\
\midrule
\cite{8942250} & Navigation and crash probability modeling & Web-based VR using real aerial data & Simulation & Crash probability, navigation accuracy & Real-world data enhances realism & No hardware validation; network latency unreported \\
\midrule
\cite{10004244} & Multi-user swarm control through VR over 5G & VR controllers + 5G testbed & Demonstration & Interaction latency, control feasibility & Low-latency multi-user control; 5G integration & Requires 5G infrastructure; scalability untested \\
\midrule
\cite{9995060} & AR with haptic feedback for collision avoidance & Wearable AR + DRL simulation & Simulation + demo & Collision rate, user anticipation & Improves human-UAV safety and interaction & Haptic latency; ergonomics issues \\
\midrule
\cite{4741347}--\cite{liu2018usability} & Pilot training and interface evaluation & VR/AR training platforms & User studies & Training time, usability scores & Effective skill transfer; reduced training time & Metrics inconsistent; motion fatigue unreported \\
\bottomrule
\end{tabularx}
\end{table*}

\subsection{Discussion}
Emerging technologies such as 5G New Radio (NR) and GAI are pivotal in advancing UAV operations and enabling XR applications. The implementation of 5G NR, developed by the 3rd Generation Partnership Project (3GPP), addresses critical requirements such as low latency, high reliability, reduced power consumption, and high capacity. These capabilities ensure seamless, immersive, and interactive XR experiences. Specifically, features like Ultra-Reliable Low Latency Communications (URLLC), Massive Machine-type Communications (mMTC), and Enhanced Mobile Broadband (eMBB) empower XR applications to handle large volumes of data with minimal delay and high uptime. For instance, 5G’s ultra-low latency supports near-instantaneous feedback between UAVs and XR environments, crucial for maintaining immersion and enhancing operational efficiency. Moreover, its high capacity facilitates simultaneous operations of multiple XR users and UAVs within the same network, enabling large-scale deployments and events in crowded environments.
\par
GAI further complements XR by automating content generation, enhancing immersive environments, and optimizing user interaction\cite{ning2026generative}. For UAV-specific applications, GAI enables dynamic, real-time simulations for training, mission planning, and predictive maintenance. It automates the creation of UAV-related content, such as flight data visualizations, which support autonomous navigation and improve operational efficiency. Additionally, AI adapts to operator behavior, offering intuitive interfaces and decision-making support, particularly in complex missions such as surveillance and rescue operations. These enhancements make UAV operations more immersive, responsive, and tailored to specific mission requirements, transforming the integration of UAVs with XR systems.
\par
However, these advancements come with significant privacy and security challenges. UAV systems collect extensive data, including telemetry, video, and sensor readings, which can be exploited by malicious actors. Common threats such as hacking, malware, ransomware, and denial-of-service attacks pose risks to UAV missions and operations. Unique vulnerabilities, such as interception of sensitive communication links or tampering with UAV controls, can result in unauthorized surveillance, data theft, or even physical harm. To mitigate these risks, robust cybersecurity measures and secure communication protocols are essential. Additionally, comprehensive reviews of UAV software and hardware systems are critical to ensuring data privacy and operational integrity. Addressing these challenges is imperative for realizing the full potential of UAVs in dynamic and secure environments.

In summary, the integration of 5G NR and GAI technologies transforms UAV operations by enabling enhanced XR applications and addressing critical operational demands. At the same time, the importance of robust security measures cannot be overstated, as they are foundational to the safe and effective deployment of these advanced technologies.

\subsection{Key Findings and Insights}

Based on the analysis in the preceding subsections, it is also evident that XR improves path planning and situational awareness and facilitates UAV control and training. In the following section, we outline key issues that require further research.

\subsubsection{Path Planning}
XR, especially AR, can significantly improve path planning by providing engineers with real-time visualizations of paths and enabling remote control for indoor tasks. In the context of Industry 4.0 and the evolving Industry 5.0, UAVs are increasingly used for indoor operations like real-time monitoring, wireless coverage, and remote sensing. Traditional path planning often faces challenges in indoor environments due to complex obstacles and confined spaces. XR technologies like AR allow engineers to design and visualize flight paths interactively, offering a more intuitive approach to path planning \cite{mourtzis2024unmanned}. By using AR, engineers can remotely design sequences of actions, view the UAV planned trajectory in real-time, and adjust the path dynamically without needing physical intervention. This approach provides enhanced flexibility, allowing the addition of multiple waypoints and real-time modifications based on the UAV surrounding environment. It also helps visualize obstacles, optimize routes, and ensure safe navigation in cluttered or confined spaces. This XR-driven method enables more efficient and precise path planning, reducing the likelihood of errors and improving operational efficiency in indoor environments, ultimately supporting better decision-making.

\subsubsection{Cybersecurity}
XR, especially MR, can significantly improve cybersecurity by providing a real-world environment to test, assess, and mitigate the impact of cyber threats such as False Data Injection (FDI) attacks [78]. By utilizing MR technologies such as Gazebo simulations and motion capture systems, XR provides a platform to emulate the physical characteristics of sensor data, including factors such as latency and noise that are often targeted by FDI attacks \cite{pant2024mixed}. This allows cybersecurity professionals to test and evaluate various attack detection algorithms and mitigation strategies in a controlled virtual environment before deploying them in real systems. By realistically replicating sensor vulnerabilities and testing mitigation strategies in an immersive environment, XR improves cybersecurity by providing a safer and more efficient way to identify, test and validate protective measures against cyberthreats.

\subsubsection{Latency}
Latency is a critical factor in XR environments, directly impacting user experience and interactivity. XR applications require ultra-low latency, typically within a few milliseconds, to provide seamless and responsive interactions, such as real-time updates in visual and aural outputs following user movements. High-speed data transmission between UAVs, edge computing nodes, and ground devices is required to realize low latency demands. The necessity of processing vast amounts of data in real-time while mitigating delays caused by factors such as network congestion, propagation delays, and computational processing gives rise to challenges that can be addressed by innovations like edge computing and advanced network management \cite{le2024reliable}, ensuring reliable, high-performance XR experiences.

\subsubsection{Ethical Challenges}

The main ethical challenges in XR technologies include data protection, privacy, and confidentiality. The concept of ethics-by-design is indeed a strong recommendation for mitigating these concerns early in the development process. Developers can proactively prevent or minimize harmful consequences by integrating ethical frameworks from the outset. However, this approach often conflicts with traditional ethics approval processes, which require a detailed ethical review before a project begins. This can create challenges for XR developers and researchers, who may face a gap between iterative design practices and the rigid, pre-project ethics review structure. To overcome this, more adaptive frameworks for ethics approval may be necessary. These frameworks would allow for continuous assessment and adaptation of ethical considerations as the project evolves, rather than requiring a one-size-fits-all solution upfront.
\subsubsection{Privacy}

XR-enabled UAV networks pose a critical privacy challenge, especially in terms of sensitive data leakage and eavesdropping. These systems generate and utilize large amounts of sensitive information, including real-time video streams, spatial mappings, and sensor data, which are essential for surveillance and disaster response applications. However, inadequate encryption or storage security mechanisms can lead to unauthorized persons accessing this data and potentially exposing classified areas. In addition, real-time communications between UAVs, ground control stations and XR interfaces are vulnerable to eavesdropping, allowing attackers to intercept key operational details such as flight paths and mission objectives. End-to-end encryption and robust storage solutions for XR-generated data are essential to eliminate these vulnerabilities. Furthermore, adopting secure communication protocols such as Transport Layer Security (TLS) or Datagram Transport Layer Security (DTLS), along with frequency-hopping techniques, can effectively mitigate these risks and enhance the security of XR-assisted UAV networks\cite{el2024cybersecurity}.

\subsubsection{Data Collection in AR and VR}

AR and VR systems are inherently data-intensive, collecting vast amounts of mission and environmental data to create immersive experiences. For example, VR headsets track UAV movements and maneuvers, while AR applications may capture real-time mission and environmental data, including a UAV’s surroundings. The importance and volume of such data exacerbate risks, including potential misuse, unauthorized sharing, or breaches.

\subsubsection{Operator Safety in AR/VR Environments}

AR and VR systems provide highly immersive experiences, closing the gap between the physical world and digital simulations. This increased immersion heightens the risk of physical injuries, such as falls or collisions, resulting from operators interacting with virtual elements in real-world UAV scenarios. Moreover, health concerns like VR-induced motion sickness or the effects of prolonged exposure to immersive environments have raised significant attention.

\section{Future Trends and Research Directions}
\label{sec4}

The preceding survey reveals both the promise and the persistent gaps in DT--XR--UAV integration. Rather than enumerating isolated improvements, we discuss below the directions we believe are most likely to yield meaningful progress.

\begin{itemize}

  \item Explainable AI (XAI).
  AI-assisted DT and XR pipelines already demonstrate strong predictive performance, yet operator trust remains a practical bottleneck. XAI~\cite{10002946} addresses this directly: by exposing the reasoning behind system recommendations~\cite{kobayashi2024explainable}, it transforms a black box into an auditable tool. Forensic traceability---the ability to reconstruct why a failure occurred---is equally important, especially in safety-critical missions where post-hoc analysis informs future design.

  \item VR Interfaces and Immersive Training.
  Two-dimensional touchscreen controls are inherently mismatched to three-dimensional flight tasks. VR interfaces~\cite{alsamhi2023flyvr} close this gap by providing spatially congruent interaction, and early evidence suggests measurable gains in both speed and accuracy for path-planning operations. The same immersive infrastructure, when configured as a training simulator~\cite{8711034}, lets operators rehearse high-risk scenarios---sensor failures, collision avoidance, forced landings---without operational consequence. Realizing this in practice demands efficient streaming of DT environmental data to headsets and robust handling of dynamic, non-static scenes.

  \item ML-Augmented Digital Twins.
  A DT that merely mirrors current state offers limited value; one that anticipates future state is operationally transformative. Integrating ML into DT pipelines enables predictive maintenance, early anomaly detection, and continuous efficiency optimization through pattern analysis across historical and live data streams. Crucially, this should not be a static deployment: a well-designed feedback loop allows models to retrain on data they themselves generate, compounding accuracy gains over time.

  \item AI, Computer Vision, and AR Convergence.
  The combination of AI inference~\cite{10208153}, computer vision-based perception, and AR overlays creates a coherent layer of augmented situational awareness that neither technology achieves alone. Object and terrain detection feed directly into AR annotations visible to the operator in real time, reducing cognitive load and accelerating reaction to hazards. For training contexts, the same pipeline provides structured, feedback-rich environments that are difficult to replicate otherwise.

  \item DT and XR for Swarm Coordination.
  Coordinating multi-UAV swarms introduces complexity that scales poorly under conventional monitoring approaches. DT replicas of individual agents and their shared environment make it possible to simulate inter-agent dependencies and stress-test coordination strategies before deployment. XR visualization then makes the resulting data legible to human operators, supporting faster intervention when swarm behavior diverges from plan.

  \item Generative AI Integration with DTs.
GAI capabilities, including large-scale data synthesis, adaptive scenario generation, and natural-language querying, map naturally onto the core functions of a DT. The prospect of asking a DT system a question in plain language and receiving a ``what-if'' analysis of mission outcomes represents a qualitative shift in how operators interact with these tools. Whether current GAI architectures are mature enough for deployment in safety-critical UAV contexts remains an open question, but the trajectory is encouraging.

  \item LLM Agents and Synthetic Data via Diffusion Models.
LLM-based agents, if integrated into DT control architectures, could autonomously manage specialized subtasks, including log analysis, anomaly triage, and configuration optimization, at a scale impractical for human operators. Separately, DMs offer a principled mechanism for generating diverse synthetic datasets to enrich DRL training, particularly when real-world data is sparse or when the underlying channel model is unknown or nonlinear, conditions under which parametric calibration methods such as RLS are insufficient. It must be noted, however, that neither direction has been empirically validated in peer-reviewed UAV DT research. Both should be treated as long-term research agendas that require systematic benchmarking and careful real-world transfer experiments before any operational claims can be made.

\end{itemize}

\section{Conclusion} \label{sec5} 

This paper provides a thorough review of current research and development efforts in the application of immersive digital technologies for UAVs. We investigate the integration of DT and XR with AI algorithms to enhance UAV systems, making them more intelligent, adaptive, and responsive. 

To ground this vision concretely, consider a UAV swarm deployed in a harsh and remote environment, such as a wildfire perimeter, a post-disaster search zone, or a remote infrastructure inspection corridor, where physical access is either unsafe or impractical. In such a setting, a DT of the entire network runs continuously, ingesting real-time telemetry from each UAV, calibrating its channel and mobility models, and maintaining a high-fidelity replica of the operational environment. Remote operators, equipped with XR headsets, connect directly to this DT and step into an immersive representation of the swarm's surroundings. From there, they can monitor individual UAV states, intervene in mission planning, and issue directives without being physically present. AI algorithms running within the DT translate these high-level operator inputs into precise flight commands, while GAI-powered interfaces allow operators to query the system in natural language and receive predictive analyses of mission outcomes. This is not a distant prospect; the building blocks reviewed in this paper, from DRL-based DT calibration and XR-assisted situational awareness to swarm coordination frameworks, are already individually demonstrated. What remains is their coherent integration. 

Realizing this vision will require advances on several fronts: tighter DT synchronization under constrained bandwidth, standardized XR interfaces for multi-operator UAV control, and robust AI pipelines that remain trustworthy under adversarial conditions. The research gaps and future directions identified throughout this paper are steps toward that goal. We hope this survey serves as a useful reference for researchers and practitioners working to bring this integrated vision to life.

\appendix[Complete Hyperparameters of Case Study I]
\label{app:hyperparams}
Table~\ref{tab:full_params} lists the complete hyperparameters of the
DT-driven DRL pipeline of Section~\ref{sec:casestudy}, covering DQN
training, the Kalman filter bank, online RLS calibration, and the
anomaly detector.

\begin{table}[h]
\centering
\caption{Complete hyperparameters for the DT-driven DRL case study (DQN training, Kalman filtering, and RLS calibration).}
\begin{tabular}{|c|c|p{3cm}|}
\hline
\textbf{Parameter} & \textbf{Value} & \textbf{Description} \\
\hline
\multicolumn{3}{|l|}{\textit{DQN Training}} \\
\hline
\texttt{max\_episodes}  & 500  & Total number of training episodes \\
\hline
\texttt{max\_timesteps} & 30   & Decision steps per episode \\
\hline
\texttt{batch\_size}    & 100  & Transitions sampled per training step \\
\hline
\texttt{learning\_rate} & $1{\times}10^{-3}$ & Adam optimiser learning rate \\
\hline
$\gamma_{\mathrm{RL}}$  & 0.99  & Discount factor in Bellman equation \\
\hline
$\tau$                  & 0.005 & Polyak coefficient for target network \\
\hline
Replay buffer           & 100{,}000 & Maximum stored transitions \\
\hline
Hidden layer width      & 256  & Units per shared backbone layer \\
\hline
\multicolumn{3}{|l|}{\textit{Kalman Filter Bank}} \\
\hline
$Q_{\mathrm{pos}}$      & 1 m$^2$   & Process noise, UAV position filters ($x$, $y$) \\
\hline
$R_{\mathrm{pos}}$      & $\sigma_p^2 = 4$ m$^2$ & Measurement noise, position filters \\
\hline
$Q_{\mathrm{pl}}$       & 0.5 dB$^2$ & Process noise, path-loss filters \\
\hline
$R_{\mathrm{pl}}$       & $\sigma_m^2 = 9$ dB$^2$ & Measurement noise, path-loss filters \\
\hline
\multicolumn{3}{|l|}{\textit{Online RLS Calibration}} \\
\hline
$\lambda_f$             & 0.98 & Forgetting factor \\
\hline
$\hat{\alpha}_0$        & 2.0  & Initial path-loss exponent estimate (vs.\ $\alpha_{\mathrm{true}}=2.8$) \\
\hline
Clip bounds             & $[1.5, 5.0]$ & Physical plausibility range enforced on $\hat{\alpha}$ after each update \\
\hline
$\Delta_{\mathrm{sync}}$ & 5 decision steps & Telemetry/synchronisation interval \\
\hline
\multicolumn{3}{|l|}{\textit{Anomaly Detection}} \\
\hline
$W$                     & 10 samples & Sliding window size for residual statistics \\
\hline
$\gamma_{\mathrm{anom}}$ & 3    & Standardised-residual (z-score) threshold \\
\hline
\end{tabular}
\label{tab:full_params}
\end{table}

\bibliographystyle{IEEEtran}
\bibliography{sn-bibliography}

\end{document}